\documentclass{article}

\usepackage{arxiv}

\usepackage[utf8]{inputenc} 
\usepackage[T1]{fontenc}    
\usepackage{hyperref}       
\usepackage{url}            
\usepackage{booktabs}       
\usepackage{amsfonts}       
\usepackage{nicefrac}       
\usepackage{microtype}      
\usepackage{lipsum}		
\usepackage{graphicx}
\usepackage{natbib}
\usepackage{doi}
\usepackage[capitalise]{cleveref}
\usepackage{subfig}
\usepackage{xcolor}
\usepackage[table]{xcolor}
\usepackage{placeins}
\usepackage{multirow}

\title{Mimir: Large-scale Multilingual Concept Modeling}

\date{} 					

\author{Elio Musacchio \\
	Department of Computer Science\\
	University of Bari Aldo Moro\\
	Bari, Italy \\
	\texttt{elio.musacchio@uniba.it} \\
	\And{Lucia Siciliani} \\
	Department of Computer Science\\
	University of Bari Aldo Moro\\
	Bari, Italy \\
	\texttt{lucia.siciliani@uniba.it} \\
	\And{Pierpaolo Basile} \\
	Department of Computer Science\\
	University of Bari Aldo Moro\\
	Bari, Italy \\
	\texttt{pierpaolo.basile@uniba.it}
}

\renewcommand{\headeright}{}
\renewcommand{\undertitle}{}
\renewcommand{\shorttitle}{}

\hypersetup{
pdftitle={Mimir: Large-scale Multilingual Concept Modeling},
pdfauthor={Elio Musacchio},
pdfkeywords={Language Modeling, Concept Modeling, Multilinguality},
}

\begin{document}
\maketitle

\begin{abstract}
    Current language modeling approaches are built around tokens. Text corpora are split into tokens, and models are trained by performing computations on these tokens, such as predicting the next token given the preceding ones as context. This paradigm has become the standard in modern language modeling, especially given the outstanding performance obtained by token-based architectures. However, recent works have not only begun to question how language models process and understand meaning from tokens, but also to question whether using higher levels of granularity could advance the research field. This led to the idea of Concept Modeling, that is, to directly train models for next-concept prediction rather than next-token prediction. 
    In this work, we introduce Mimir, a 1.6B Large Concept Model trained for multilingual concept understanding and generation. We leverage a large-scale multilingual pre-training corpus (38,883,987,240 sentences) spanning 46 languages and a large-scale multi-turn, and multilingual instruction-tuning dataset (66,816,428 sentences) covering a total of 35 languages. We extensively evaluate model performance against a language model with a comparable number of parameters.
\end{abstract}


\section{Introduction}
\label{sec:intro}

In recent years, Large Language Models (LLMs) have completely revolutionized the field of Natural Language Processing (NLP) \citep{qin2026large}. 
These models work on massive text corpora, split into tokens, and learn to predict the next token autoregressively based on the tokens provided as context. 
Despite the undeniable success of this paradigm, recent work has begun to explore alternatives to traditional token-level modeling.

One emerging research direction is \textbf{Concept Modeling}, as introduced by Large Concept Models \citep{lcm} (LCMs). LCMs treat sentences as concepts and perform the autoregressive prediction objective on concepts rather than tokens \citep{beyondtokens}. This approach encourages the model to reason at a higher level rather than relying solely on fine-grained lexical patterns.

One of the most remarkable advantages of the approach proposed by \citet{lcm} is that the model is inherently capable of multilingual generation.
This happens because concepts are represented as sentence embeddings, which can be decoded into different languages through a multilingual decoder \citep{artetxe2019massively}.
However, as with token-based LLMs, training LCMs on data focused on a single language (e.g., English) limits their ability to support multilingual understanding.
Current works on Concept Modeling have focused almost exclusively on English, without considering other non-English languages. This leaves a research gap in multilingual concept modeling.
In light of this, we propose Mimir, the first Large Concept Model trained on large-scale multilingual data.

Hence, the contributions of this work are the following:
\begin{itemize}
    \item We propose Mimir, the first Large Concept Model trained on multilingual large-scale data. We propose a 1.6B model trained on a corpus consisting of 38,883,987,240 multilingual sentences covering 46 languages;
    \item We perform evaluation on several languages and provide a comparison with a Large Language Model within a comparable range of parameters.
\end{itemize}

We provide all resources associated with this work to facilitate reproducibility and boost the current research trends in Concept Modeling\footnote{\url{https://huggingface.co/collections/mimir-lcm/mimir}}.

\section{Related Works}
\label{sec:related_works}

\subsection{Concepts and LLMs}

Several recent works have studied the relationship between LLMs and concepts. 
Early studies focused on understanding whether LLMs encode and manipulate conceptual knowledge effectively.
In \citet{peng2022copen}, the authors addressed the lack of benchmarks targeting conceptual rather than factual knowledge. 
To overcome this limitation, they collected 24,000 instances spanning 393 concepts and conducted extensive experiments on pre-trained language models, revealing that these models lacked conceptual knowledge.
Other works have begun to question how concepts are internally represented within an LLM and whether a paradigm shift is possible. 
For example, \citet{jin2025exploring} introduced the idea of "Concept Depth". 
The authors suggested that LLMs learn concepts of varying difficulty across different layers, with more complex concepts learned at deeper layers of the model.
Similarly, \citet{han-etal-2025-towards} presented a concept editing method for LLMs. 
Concept editing approaches focus on modifying the representations of specific concepts within LLMs to guide their outputs. 
They proposed a unified neuron-level paradigm for concept editing, a common framework for understanding and comparing diverse editing methods.
\citet{bhan-etal-2025-towards} proposed the Complete Textual Concept Bottleneck Model, a method for generating concept labels that leverages a small, fine-tuned classifier language model.

Beyond analyzing conceptual representations in token-based LLMs, recent work has begun to rethink the language modeling objective itself. 
\citet{beyondtokens} proposed replacing the traditional next-token prediction objective with next-concept prediction, demonstrating that next-concept prediction performs better in terms of perplexity w.r.t. the traditional next-token prediction objective used in LLMs.
\citet{lcm} proposed the Large Concept Model, a model to perform autoregressive concept prediction.
Concepts are treated as sentences and encoded in the SONAR \citep{sonar} embedding space.
They propose three variants of the LCM model: Base, One-Tower, and Two-Tower.

Despite the efforts proposed to bridge the gap between concepts and LLMs, very few studies consider implementing a Large Concept Model and mostly focus on evaluating pre-trained LLMs with respect to their conceptual knowledge.

\subsection{Multilingual LLMs}

In recent years, there has been increasing interest in multilingual LLMs, with research focusing on scaling, evaluation, and cross-lingual learning.
For example, \citet{tanwar-etal-2023-multilingual} explored in-context learning in a cross-lingual setting. 
They proposed a prompt construction strategy to replace the random selection of labeled training examples by combining semantic and task-based alignment. 
Subsequent efforts have focused on scaling the multilingual capabilities of these models to a larger number of languages.
\citet{lai-etal-2024-llms} introduced xLLaMA-100 and xBLOOM-100, extending multilingual support to 100 languages.
To do so, they proposed two datasets: one covering 100 languages for multilingual instruction tuning and one covering 30 languages for cross-lingual human preference modeling.
Also focusing on scaling, \citet{geigle-etal-2025-centurio} presented an exhaustive study on training strategies for massively multilingual multimodal LLMs. 
They studied best practices for multilingual and multimodal training data and proposed the Centurio model, a multimodal LLM supporting 100 languages.

Alongside model scaling, several works have addressed the evaluation and analysis of multilingual semantic capabilities.
\citet{ying-etal-2025-disentangling} introduced a Dual Evaluation Framework for assessing the multilingual capabilities of LLMs. 
Focusing instead on internal semantic representations, \citet{korner-etal-2026-meanings} studied the development of language-agnostic concept spaces during pre-training of EuroLLM  \citep{martins2025eurollm}.
EuroLLM is a suite of open-weight multilingual LLMs capable of understanding and generating text in all official European Union languages.
They showed that shared concept spaces emerge early and remain relatively stable throughout training.

Furthermore, multilingual research also expanded towards cultural understanding. \citet{nyandwi-etal-2025-grounding} introduced CulturalGround, a dataset for evaluating multimodal LLMs' cultural understanding across 42 countries and 39 languages.
They also introduced CulturalPangea, a multilingual and multimodal LLM that achieves state-of-the-art on culture-focused benchmarks.

Despite the rapid progress in multilingual language modeling, existing work remains largely intertwined with token-based architectures. To the best of our knowledge, no prior work has yet proposed a Large Concept Model trained on large-scale multilingual data.

\section{Data Collection}
\label{sec:data}

\begin{figure*}[ht]
  \centering
   \includegraphics[width=0.8\textwidth]{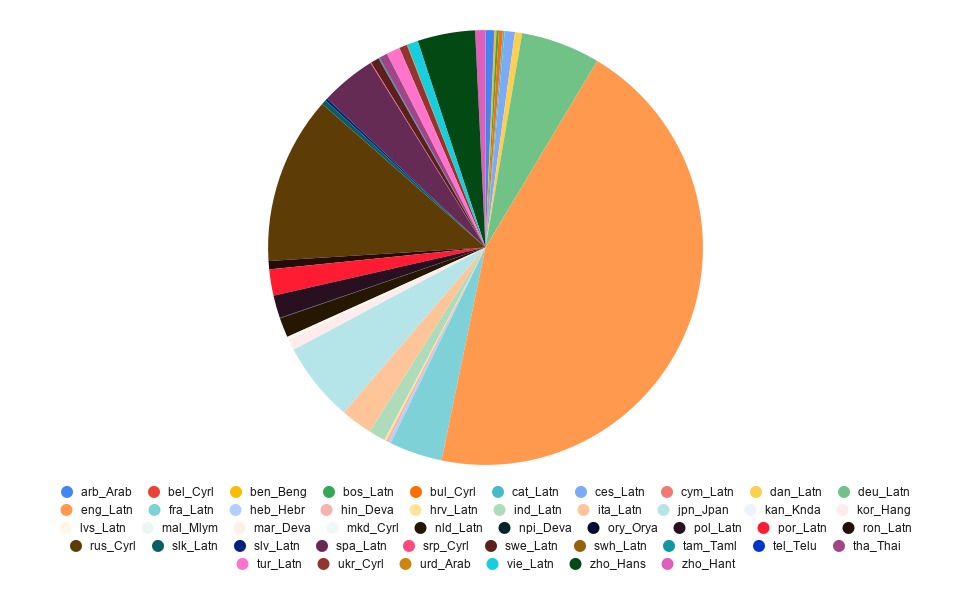}
   \caption{Pie chart of the distribution of sentences for the pre-training dataset languages}
   \label{fig:pie}
\end{figure*}

To train the model, we collected large-scale pre-training multilingual data.
Specifically, for pre-training, we use the 350BT split of the Fineweb-edu dataset \citep{lozhkov2024fineweb-edu} for the English language and the Fineweb 2 dataset \citep{penedo2025fineweb2pipelinescale} for other languages.
Fineweb-edu consists of educational web pages extracted from the Fineweb dataset, while Fineweb 2 is the second iteration of the dataset, comprising high-quality pre-training data for more than 1,000 languages.
For Fineweb 2, we sample data for the languages so that the pre-training corpus is equally split into Fineweb-edu and Fineweb 2.

An overview of the language distribution in the pre-training dataset is shown in \Cref{fig:pie}, while detailed statistics are reported in \Cref{sec:appendix_data}.
We also note that Fineweb 2 unifies Chinese data under a unique language code, namely ``cmn\_Hani'' (Mandarin Chinese). 
We distinguish between simplified (``zho\_Hans'') and traditional (``zho\_Hant'') Chinese using the \textit{hanzidentifier}\footnote{\url{https://github.com/tsroten/hanzidentifier}} library.
We perform this additional processing step because SONAR expects either ``zho\_Hans'' or ``zho\_Hant''.
Overall, across the entire pre-training dataset, we report 38,883,987,240 sentences covering 46 languages, of which 44.65\% are in English and 55.35\% in non-English languages.

\begin{figure*}[ht]
  \centering
   \includegraphics[width=0.8\textwidth]{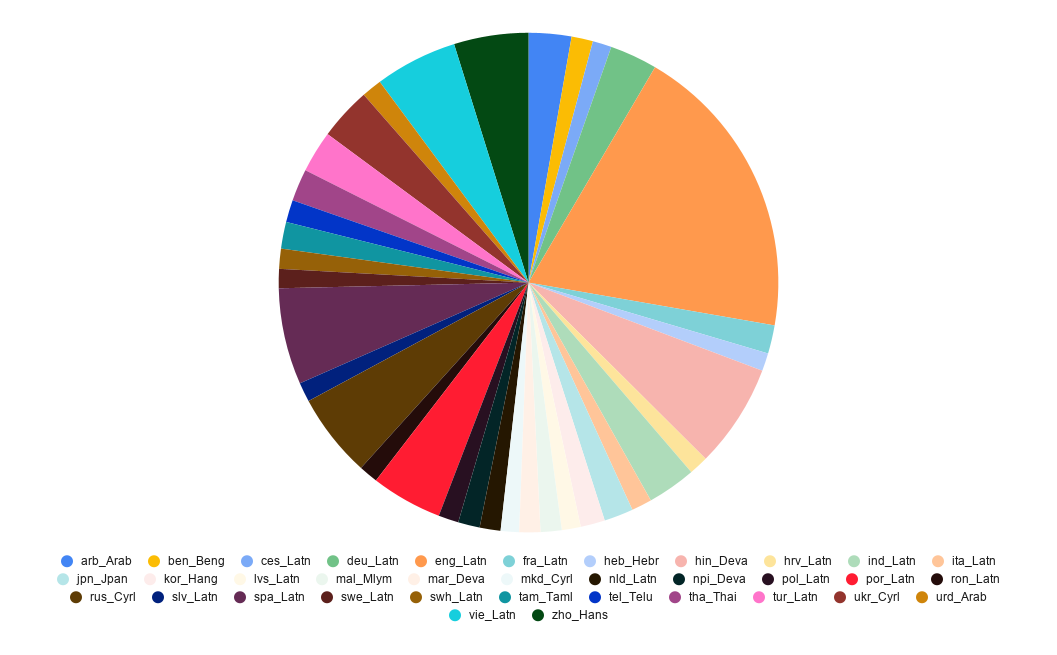}
   \caption{Pie chart of the distribution of sentences for the instruction-tuning dataset languages}
   \label{fig:pie_inst}
\end{figure*}

For our instruction-tuning datasets, we consider the following datasets: 1) \textbf{Aya Dataset} \citep{singh2024aya}, a multilingual instruction-tuning dataset curated by humans through an annotation platform. The dataset contains 204k human-annotated prompt-completion pairs across 65 diverse languages; 2) \textbf{Bactrian-X} \citep{li2023bactrianx}, a collection of 3.4M instruction-response pairs in 52 languages that are obtained by translating 67k English instructions into 51 languages using the Google Translate API. The translated instructions are then given to GPT-3.5-Turbo to obtain more natural responses; 3) \textbf{Open Assistant 2}\footnote{\url{https://huggingface.co/datasets/OpenAssistant/oasst2}}, a collection of conversations collected from the open-assistant.io website. Since each conversation can have multiple paths, we reconstruct the best conversation based on the top-ranked response. The final dataset we processed consists of 19,654 instances covering 20 languages.
We also added a synthetically generated multilingual dataset curated by us.
Our objective was to create a multilingual multi-turn data mixture with culturally sensitive elements.
We prompt the Qwen3-235B-A22B-Thinking-2507 model to provide conversational turns between a user and an AI assistant based on a given topic.
The topics are sampled from the Everyday Conversations dataset \citep{everydayconversations2024}, which contains 2,200 multi-turn conversations generated by Llama-3.1-70B-Instruct on a given topic.
The prompt used to generate this data is shown in \Cref{sec:appendix_synth}.

We also add a multilingual instruction-tuning dataset focused on math subjects to improve model performance on math-related tasks.
In this case, we extract math problems and their solutions from the OpenMathInstruct-2 dataset \citep{toshniwal2024openmath2}, a math instruction-tuning dataset comprising 14M problem-solution pairs generated with the Llama-3.1-405B-Instruct model.
The model is prompted to provide a conversation for that math problem in the target language and continue the conversation for additional turns.
The prompt used to generate this data is shown in \Cref{sec:appendix_synth}.

For both Everyday Conversations and OpenMathInstruct, we include instructions in the prompt to randomize and diversify conversations (e.g., random number of turns, answer concisely).

Finally, we also include the train sets from MLQA \citep{lewis2020mlqa} and XL-Sum \citep{hasan2021xl} in the instruction-tuning mixture.
XL-Sum is a comprehensive and diverse dataset comprising 1.35M professionally annotated article-summary pairs from BBC, extracted using a set of carefully designed heuristics, and covering 44 languages. 
MLQA is an extractive question-answering dataset over paragraphs covering 7 languages.
The MLQA train set is derived by machine translating the SQuAD dataset \citep{rajpurkar-etal-2016-squad} into 7 languages (i.e., English, Arabic, German, Spanish, Hindi, Vietnamese, and Simplified Chinese).

To study both multilingual generalization and the impact of task-specific supervision, we consider four instruction-tuning settings: 1) multilingual with MLQA and XL-Sum in the train set; 2) multilingual without MLQA and XL-Sum in the train set; 3) English only with MLQA and XL-Sum in the train set; 4) English only without MLQA and XL-Sum in the train set.
We consider variants with and without MLQA and XL-Sum in the training set to evaluate Mimir's generalization capabilities on tasks not seen during training.
Similarly, the English-only variant has the aim to assess the performance discrepancy when using a model instruction-tuned only on English.

An overview of the instruction-tuning dataset cardinality per language is shown in \Cref{fig:pie_inst}, while the complete statistics are reported in \Cref{sec:appendix_data}.
Overall, across the whole instruction-tuning dataset, we report 66,816,428 sentences covering 35 languages, of which 19.29\% are in English and 80.71\% in non-English languages.

In all cases, we split data into sentences using the \textbf{sat-3l} \citep{frohmann-etal-2024-segment} model from \textit{wtpsplit}\footnote{\url{https://github.com/segment-any-text/wtpsplit}} \citep{minixhofer-etal-2023-wheres}. 
We fix a sentence threshold of 0.02 and a maximum sentence length of 256.

\section{Methodology}
\label{sec:method}

We follow the implementation of the Two-Tower diffusion LCM model by \citet{lcm}. 
In this framework, concepts are treated as sentences, and the textual corpora are processed by splitting the input into distinct sentences.
The architecture consists of two components: 1) a context encoder, where a decoder-only transformer is used to encode the contextual sequence of previous sentence embeddings; 2) a denoiser, where a stack of transformer blocks with cross-attention is used to attend over the encoded context representations.
Hence, the model training objective is to generate the SONAR embedding of the next concept, given the previous sentences as context.
Then, to generate text, a decoder reconstructs the original sentence from the embedding.
Both the embedding encoder and the decoder are inherited from SONAR \citep{sonar}.

One advantage is their inherent support for multilingual language generation.
Since the SONAR decoder can reconstruct text in any supported target language, even if the input embedding is encoded in English, it can be decoded into another language.
However, while this solves the problem of multilingual generation, it does not address multilingual language understanding. 
While the SONAR decoder can reconstruct text in any language, the denoiser and context encoder are trained only on English embeddings.
Moreover, SONAR embeddings are not aligned with language, implying that semantically equivalent sentences across different languages may have different embeddings. 
To investigate this limitation, we conduct a pilot experiment to measure the alignment between English and non-English SONAR embeddings in a parallel corpus.
Specifically, we use the ``parallel-sentences-wikimatrix''\footnote{\url{https://huggingface.co/datasets/sentence-transformers/parallel-sentences-wikimatrix}} dataset as a parallel corpus.
We first extract the first 1,000 sentences for each language subset from the ``dev'' split, and then compute the cosine similarity between the English sentence embedding and its corresponding non-English embedding.
The average cosine similarity is computed for all 1,000 pairs.
The results are shown in \Cref{sec:appendix_eval}.
Overall, the results show that SONAR embeddings are not perfectly aligned across languages, with cosine similarity scores below 0.90 for all subsets. 
This implies that models trained exclusively on English embeddings may develop a bias towards this language, making them less effective at handling prompts in other languages.
A visualization of the differences between the two approaches is shown in \Cref{sec:appendix_method}.
In light of this, we pre-train and fine-tune an LCM model on large-scale multilingual data.

For pre-training, the model is trained for 250,000 steps, with a 4e-4 learning rate, a cosine scheduler, 10,000 warmup steps, 0.1 weight decay, and a batch size of 229,376 sentence embeddings, following the original LCM model recipe.
For instruction tuning, we train for 20,000 steps with a 1e-5 learning rate, a cosine scheduler, 0.01 weight decay, and a batch size of 512 instances.
Rather than the autoregressive token-prediction objective, LCM models are trained on an autoregressive concept-prediction objective.
The next-concept prediction objective is implemented using the MSE loss between the embedding generated by the model and the ground truth embedding of the input sentence in the SONAR space. 

For pre-training, the entire input embeddings are processed autoregressively.
Additionally, we append an ``End of text.'' sentence at the end of all inputs.
We use this sentence to stop generation at inference time when the model generates an embedding with a cosine similarity of 0.90 or greater to the SONAR embedding of the ``End of text.'' sentence.
Furthermore, following the original recipe by \citet{lcm}, we also trained a normalizer on the same data used for pre-training.

For instruction tuning, we adopt a formulation similar to completion-only training in traditional LLMs.
More specifically, there is a fixed context not used for prediction (source embeddings) while the model is trained to predict only some fixed target concepts (target embeddings).
In our setting, the user turn represents the source embeddings, while the assistant turn represents the target to be predicted. 
For multi-turn data, we consider all previous user-assistant turns as context, alongside the current user turn.
Hence, multi-turn data are split into multiple instances according to the number of turns (e.g., if a conversation has 4 turns, 4 instances are added to the dataset).
We separate the user and assistant turns using dedicated sentences, namely ``User turn.'' and ``Assistant turn.''.
Finally, we append the ``End of text.'' sentence at the end of the assistant turn to stop generation during inference using the same strategy explained in pre-training.
A formatting example is shown in \Cref{sec:appendix_formatting}.

We perform pre-training and instruction tuning on a cluster of A100 GPUs with 64GB of VRAM, using 4 nodes, each with 4 GPUs.
Pre-training took approximately a month, while instruction tuning took about 10 hours.
Unlike the original LCM implementation \citep{lcm}, we extract embeddings at runtime due to space limitations of the cluster where we perform the experiments.

\section{Experiments}
\label{sec:experiments}

We perform inference by leveraging the same parameters used by \citet{lcm}. Specifically, we use 40 inference timesteps, an initial noise scale of 0.6, a guidance scale of 3.0, a guidance rescale of 0.7, and an epsilon scaling of 1.00045.
Additionally, for pre-training evaluation, we limit generation to a single sentence, while for instruction-tuning evaluation, we limit generation to 16 sentences.
Generated embeddings are always decoded into text using the original SONAR decoder.
To assess the effectiveness of instruction tuning, we compare Mimir with Qwen3 1.7B \citep{qwen3technicalreport}, a modern LLM within the same parameter range.
For Qwen3 1.7B, we use greedy decoding directly during inference.
We do not evaluate the LCM model \citep{lcm} since the checkpoint was never publicly released.
Furthermore, we do not compare Mimir to baselines with English-only pre-training because there are no publicly available LCM baselines trained on English-only data.
For pre-training, we evaluate on three large-scale multilingual datasets: 1) \textbf{C4}\footnote{\url{https://huggingface.co/datasets/allenai/c4}} \citep{raffel2020exploring}, a cleaned version of Common Crawl's web crawl corpus; 2) \textbf{MultiEURLEX} \citep{chalkidis-etal-2021-multieurlex}, a dataset consisting of 65k European laws in 23 official European languages; 3) \textbf{Wiki40B} \citep{guo2020wiki}, a processed version of Wikipedia consisting of the full Wikipedia article after page processing that removes non-content sections and structured objects.
For pre-training evaluation, we consider the \textbf{L2 distance} (L2), the \textbf{Round-trip L2 distance} (R-L2), the \textbf{Paraphrasing} (PAR) and the \textbf{Contrastive Accuracy} (CA) metrics \citep{lcm}.
L2 distance is the Euclidean distance between the predicted embedding and the ground truth embedding.
Round-trip L2 distance is the Euclidean distance between the embedding of the decoded sentence (re-encoded in the SONAR embedding space) and the ground truth embedding.
Paraphrasing is the maximum cosine similarity between the generated embedding and the context embeddings, normalized by the maximum cosine similarity between the ground truth embedding and the context embeddings (essentially, it measures the degree to which the model relies on copying or rephrasing previous sentences w.r.t. the ground truth).
Contrastive Accuracy is the ratio of times the generated embedding is closer (in terms of L2 distance) to the actual ground truth embedding than it is to other embeddings from different sentences in the same batch.

These metrics cannot be used with LLMs, since they require a model capable of generating a sentence embedding.
Similarly, metrics used to evaluate pre-training quality for LLMs (e.g., perplexity) cannot be used to evaluate LCMs, since they do not produce a probability distribution over the next token.
Hence, we focus on evaluating Mimir's pre-training quality directly, without comparing it to Qwen.

We extract the first 1,000 instances with at least 9 sentences for each dataset and language. 
Then, we perform inference for each sentence within each instance.
That is, we perform inference by providing the model only the first sentence as context, then the first and second sentences as context, and so on.

For instruction tuning, we evaluate on two multilingual generative benchmarks: 1) XL-Sum \citep{hasan2021xl}; 2) MLQA \citep{lewis2020mlqa}. 
These datasets are described in \Cref{sec:data}, and their training set is included in our main instruction-tuning mixture.
We also perform contamination analysis of the training and test sets of these two datasets and report the results in \Cref{sec:appendix_contam}.
For instruction-tuning evaluation, we consider the \textbf{ROUGE-L} metric, following \citet{lcm}.

For pre-training, we perform evaluation on a subset of languages that are: 1) supported by both C4 and Wiki40B; 2) supported by SONAR.
For instruction tuning, we perform evaluation only on the language subsets present in the training mixture.

\subsection{Pre-Train Results}

\begin{table*}[htb]
  \centering
  \footnotesize
  \begin{tabular}{@{}lllll@{}}
    \toprule
    \textbf{Dataset} & \textbf{L2} & \textbf{R-L2} & \textbf{PAR} & \textbf{CA} \\
    \midrule
    C4 & 0.2551 & 0.2411 & 2.4839 & 72.54\% \\
    MultiEURLEX & 0.2063 & 0.1972 & 1.7796 & 56.65\% \\
    Wiki40B & 0.2980 & 0.2793 & 2.3714 & 77.70\% \\
    \bottomrule
  \end{tabular}
  \caption{Results of pre-train evaluation}
  \label{tab:pretrain_eval}
\end{table*}

\begin{figure*}[ht]
  \centering
   \includegraphics[width=0.9\textwidth]{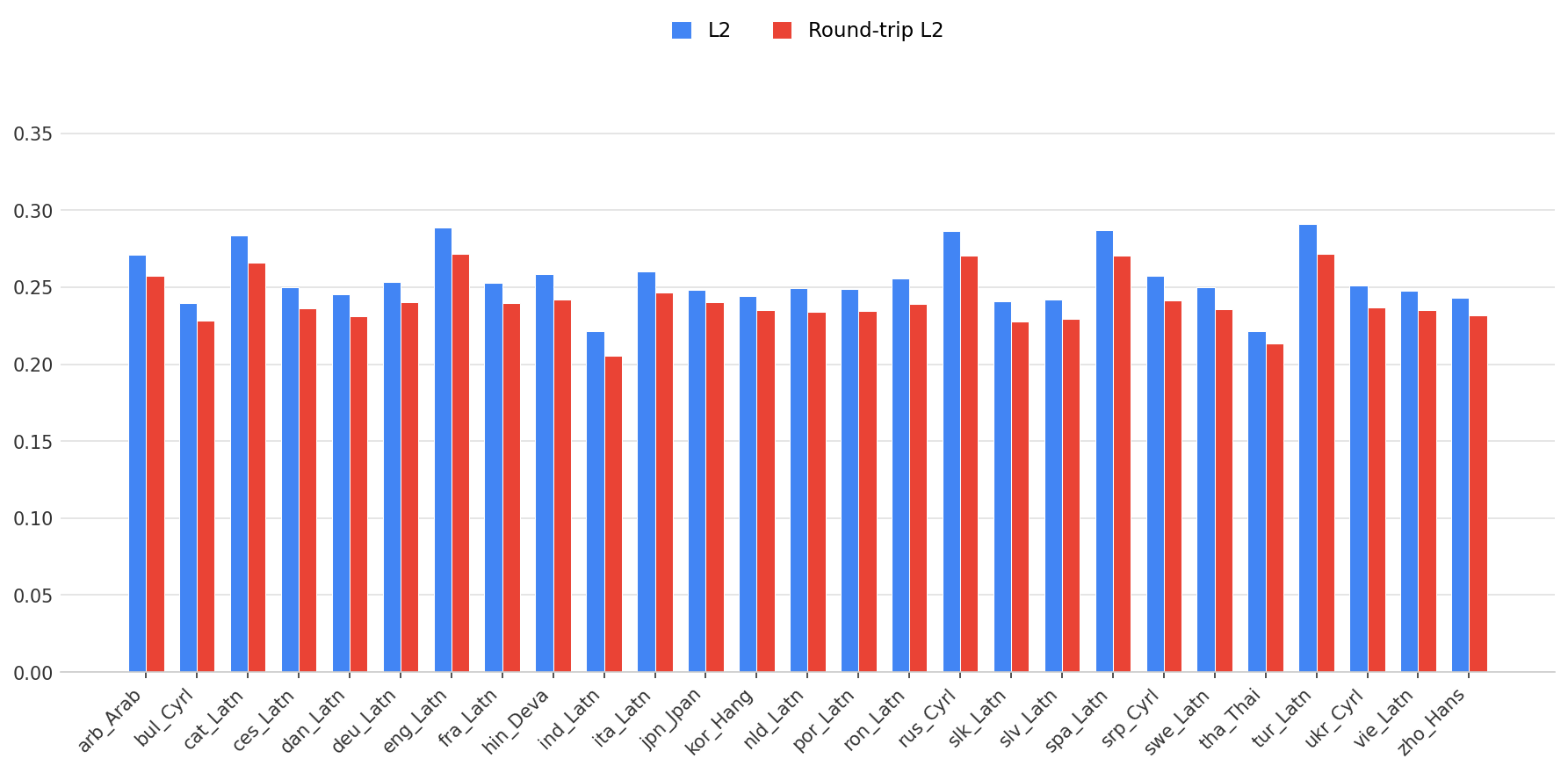}
   \caption{Bar chart showcasing pre-train evaluation results using L2 and Round-trip L2 for the C4 dataset}
   \label{fig:c4_language_results}
\end{figure*}

We report average pre-training evaluation results for C4, MultiEURLEX, and Wiki40B in \Cref{tab:pretrain_eval}.
Per-language results for the C4 dataset on L2 and R-L2 are shown in \Cref{fig:c4_language_results}, while the results for Wiki40B and MultiEURLEX on L2 and R-L2, as well as the results for CA and PAR, are shown in \Cref{sec:appendix_eval}.

Overall, the results consistently show that R-L2 is lower than L2, implying that the decoder can produce coherent sentences from embeddings.
From the average results, we find that Mimir performs best on MultiEURLEX, except for CA.
This is expected since legal documents use highly constrained and repetitive vocabulary; hence, the model can easily predict the exact continuation. 
However, this same homogeneity makes the CA evaluation unfair, because other sentences in a batch of legal texts are likely to be semantically similar to the ground truth.
Furthermore, we find similar patterns for both C4 and Wiki40B, which exhibit higher PAR w.r.t. MultiEURLEX (since they are open-domain corpora).
Interestingly, Mimir performs best on Wiki40B in terms of CA, but worse in terms of L2 and R-L2 w.r.t. C4.
We attribute this to the rigorously cleaned structure of Wikipedia documents (for CA) and to the diverse vocabulary of Wikipedia text (for L2 and R-L2).

From the language split results, we find that Mimir performs worse in English than in other languages across all three datasets, particularly for L2 and R-L2.
We attribute this to the nature of the pre-training data, given that we extracted all English data from Fineweb-edu, while we used Fineweb 2 for the other languages.
Importantly, the results for both L2 and Round-trip L2 are comparable to those obtained by \citet{lcm} on their pre-training evaluation benchmarks.
This demonstrates that the model can understand context and generate embeddings that are natural continuations of it.
Finally, manual inspection of the generated outputs further confirms that the model produces semantically consistent continuations across languages.

\subsection{Instruction Tuning Results}

\begin{table*}[htb]
  \centering
  \footnotesize
  \begin{tabular}{@{}lllll@{}}
    \toprule
    \textbf{Model} & \textbf{EN} & \textbf{OOD} & \textbf{Dataset} & \textbf{ROUGE-L} \\
    \midrule
    Mimir 1.6B & X & X & MLQA & 36.12 \\
    Mimir 1.6B & X & X & XL-Sum & 20.73 \\
    \midrule
    Mimir 1.6B & X & \checkmark & MLQA & 18.64 \\
    Mimir 1.6B & X & \checkmark & XL-Sum & 12.85 \\
    \midrule
    Mimir 1.6B & \checkmark & X & MLQA & 34.82 \\
    Mimir 1.6B & \checkmark & X & XL-Sum & 20.68 \\
    \midrule
    Mimir 1.6B & \checkmark & \checkmark & MLQA & 16.11 \\
    Mimir 1.6B & \checkmark & \checkmark & XL-Sum & 11.49 \\
    \midrule
    Qwen3 1.7B & - & - & MLQA & 36.71 \\
    Qwen3 1.7B & - & - & XL-Sum & 10.97 \\
    \bottomrule
  \end{tabular}
  \caption{Results for evaluation on MLQA and XL-Sum}
  \label{tab:it_eval}
\end{table*}

\begin{figure*}[ht]
  \centering
   \includegraphics[width=0.9\textwidth]{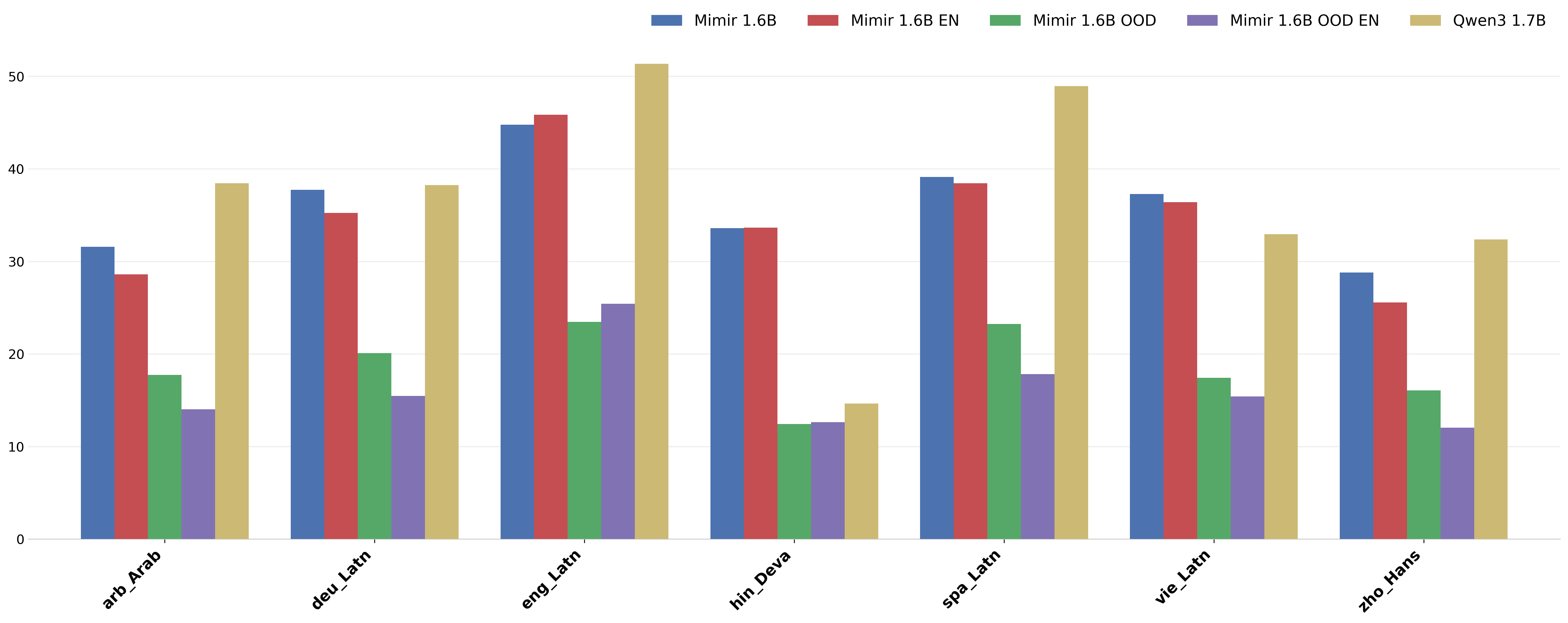}
   \caption{Bar chart showcasing instruction-tuning evaluation results using ROUGE-L for the MLQA dataset}
   \label{fig:mlqa_language_results}
\end{figure*}

\begin{figure*}[ht]
  \centering
   \includegraphics[width=0.9\textwidth]{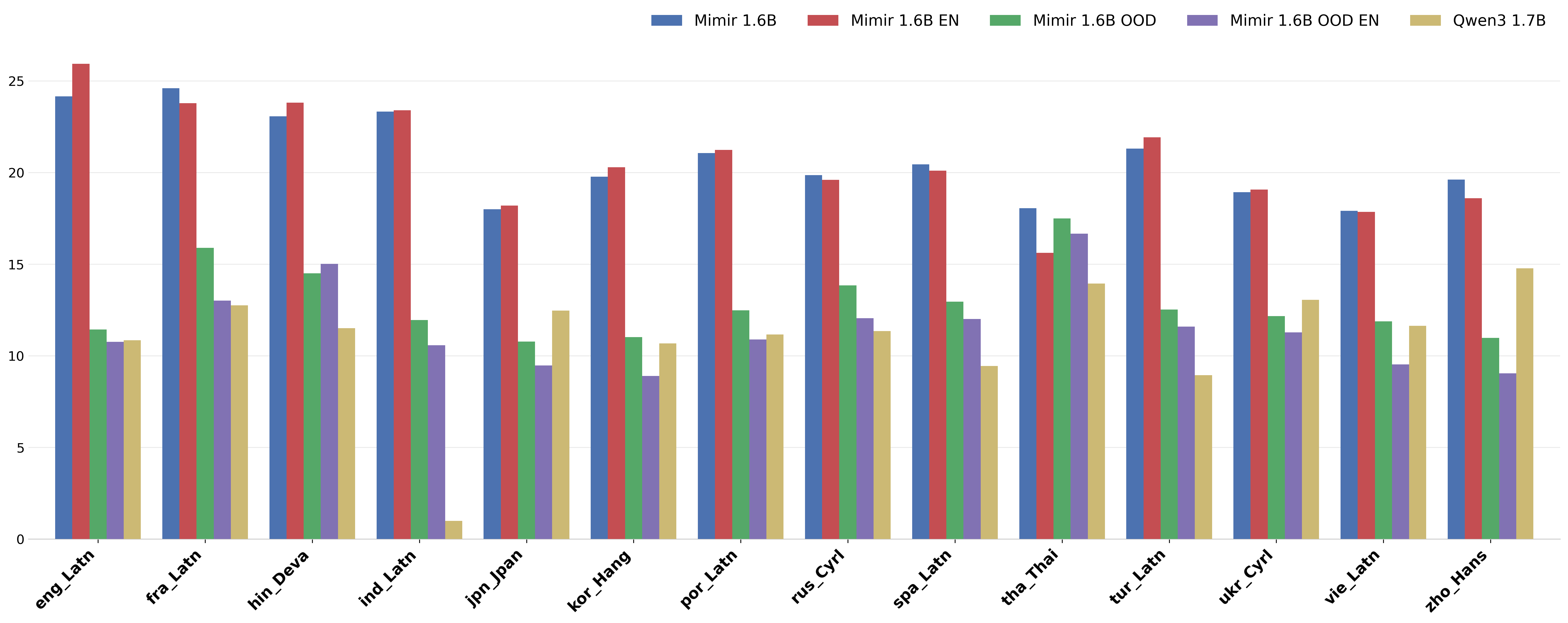}
   \caption{Bar chart showcasing instruction-tuning evaluation results using ROUGE-L for the XL-Sum dataset}
   \label{fig:xlsum_language_results}
\end{figure*}

We report average results in \Cref{tab:it_eval}, while per-language results for MLQA and XL-Sum are shown in \Cref{fig:mlqa_language_results} and \Cref{fig:xlsum_language_results}, respectively.
Configurations with English only in the train set are referred to as EN, while configurations without MLQA and XL-Sum in the train set are referred to as OOD.

Overall, Mimir performs better on XL-Sum, even when XL-Sum is not included in the train set.
Remarkably, the model trained with XL-Sum outperforms Qwen3 significantly across all languages for XL-Sum.
We find that performance is slightly lower on MLQA when compared with Qwen3, suggesting that the model is currently better suited to summarization than to question-answering tasks.
Finally, we observe that the model trained on multilingual data performs best on MLQA, while models trained with MLQA and XL-Sum in the train set consistently outperform those trained without them.
Still, models trained without MLQA and XL-Sum show a good degree of generalizability w.r.t. Qwen3 (e.g., Mimir performs better on ``fra\_Latn'' w.r.t. Qwen3 on XL-Sum).

After manually validating the outputs, we find that the model trained without MLQA and XL-Sum in the training set produced significantly worse-quality responses.
Specifically, we found the following limitations: 1) task drift, where the model was unable to provide a response relevant to the task (e.g., generating a new question instead of answering the input question in MLQA); 2) text repetition, where the model repeated the same terms within the same sentence (e.g., ``\textit{Walter Kasza and Walter Kasza}'').
We attribute these limitations to two main factors. 
The first concerns model size: since the current model has 1.6B parameters and the next-concept prediction task is significantly more complex than next-token prediction, improved performance is expected with a larger parameter count. 
Secondly, without MLQA and XL-Sum in the train set, the model was never directly exposed to extractive question-answering and summarization data.

Furthermore, we report the results for additional LLM baselines in \Cref{sec:appendix_eval}, providing broader context for the evaluation.

\section{Conclusions and Future Works}
\label{sec:conclusions}
In this work, we have introduced Mimir, the first Large Concept Model trained on large-scale multilingual data. 
We pre-train the model on a multilingual corpus of 38,883,987,240 sentences, and we perform instruction tuning on a multilingual dataset of 66,816,428 sentences.
Through extensive pre-training and instruction tuning, we investigated the capabilities and limitations of multilingual concept-level language modeling across four instruction-tuning mixtures. 
We show that Mimir performs optimally on long-context tasks (e.g., XL-Sum) and outperforms Qwen3 1.7B across most non-English languages. 
This result suggests that concept-level modeling is a promising research direction for multilingual semantic content generation.
At the same time, our analysis revealed the model's limitations, mainly related to difficulties in task generalization, task drift, and repetitive generation patterns.
These challenges indicate that concept-level language modeling remains significantly more demanding than traditional token-based language modeling, especially with smaller parameter counts.
As future work, we are developing a 7B version of Mimir and plan to extend both models to multimodal inputs. 
We additionally aim to extend both training and evaluation settings by including multiple-choice reasoning datasets and more challenging multilingual understanding tasks.

\section*{Limitations}

Our work has the following limitations:
\begin{itemize}
    \item \textbf{Embedding Space}: our work relies on SONAR \citep{sonar} as embedding model following the work of \citet{lcm}. We do not propose a study on the applicability of other embedding models to train a Large Concept Model. 
    \item \textbf{Evaluation Protocol}: currently, we evaluated only on MLQA and XL-Sum to cover a wide range of languages. However, Mimir should be evaluated on several additional benchmarks that cover diverse skills, such as multi-turn conversations and math. We plan to further extend the Mimir evaluation. Furthermore, our current experiments do not allow us to identify the underlying cause behind the lower performance w.r.t. token-level models, and future work on this topic is needed (e.g., studying the sensitivity to the backbone embedding model).
    \item \textbf{Human Assessment}: evaluation of Mimir outputs by human evaluators 
    would have required access to qualified annotators for each language as well as substantial costs in terms of time and resources. This goes beyond the scope of this work; nevertheless, it remains an important future objective that can strengthen the conclusions and the evaluation's soundness.
\end{itemize}

\section*{Ethical Considerations}

In this work, the model was pre-trained using Fineweb-edu and Fineweb 2.
While these datasets underwent rigorous quality checks and filtering by their original creators, we used them in their released state without introducing any secondary processing pipelines.

\section*{Acknowledgments}

We acknowledge the support of the PNRR project FAIR - Future AI Research (PE00000013), Spoke 6 - Symbiotic AI (CUP H97G22000210007) under the NRRP MUR program funded by the NextGenerationEU. 
We acknowledge the CINECA award under the ISCRA initiative for the availability of high-performance computing resources and support.

\appendix
\crefalias{section}{appendix}

\section{Data Details}
\label{sec:appendix_data}

Complete statistics are reported in \Cref{tab:dataset} for the pre-training dataset and in \Cref{tab:dataset_it} for the instruction-tuning dataset.

\section{Formatting}
\label{sec:appendix_formatting}

We report the formatting used for instruction tuning in \Cref{tab:prompt}. 

\section{Methodology}
\label{sec:appendix_method}

\begin{figure*}[ht]
  \centering
   \includegraphics[width=\textwidth]{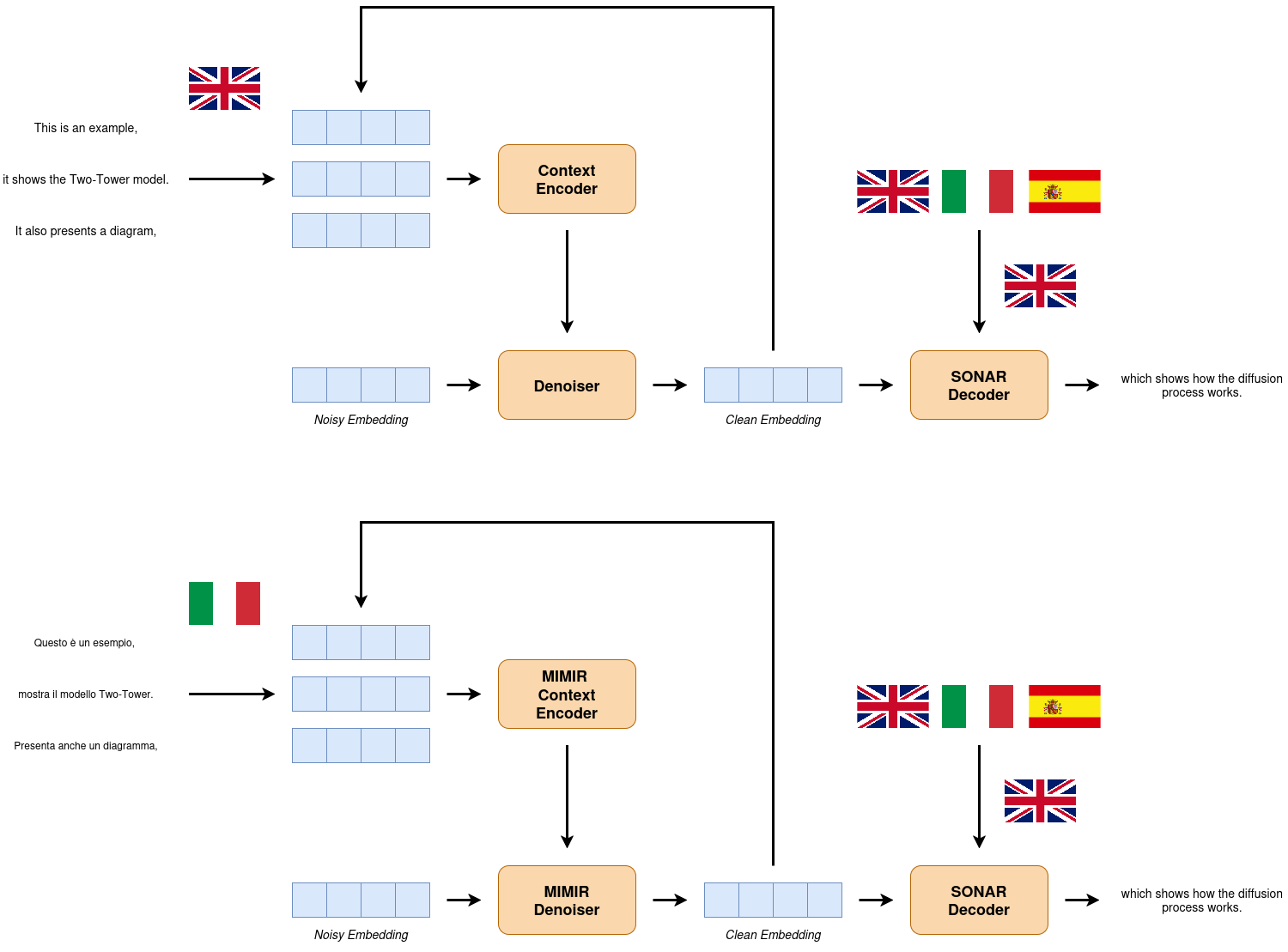}
   \caption{Overview of the diffusion process in the Two-Tower LCM architecture. The context encoder encodes SONAR sentence embeddings in English, which are then processed by the denoiser via cross-attention. The denoiser generates a clean embedding by attending the context. This clean embedding can then be used by the SONAR decoder to reconstruct text in any target language it supports. The cleaned embedding is also concatenated with the input embeddings to continue generating the next sentence. While the original LCM model only allows encoding of English sentences, Mimir allows encoding of sentences in any language and still supports generation in any language supported by the decoder.}
   \label{fig:diffusion}
\end{figure*}

We present a diagram representing Mimir and the methodological differences w.r.t. the Two-Tower LCM \citep{lcm} in \Cref{fig:diffusion}.

\section{Data Generation}
\label{sec:appendix_synth}

We report complete prompts used for the multilingual multi-turn dataset in \Cref{tab:prompt_synth} and for the multilingual multi-turn math dataset in \Cref{tab:prompt_synth_math}.

\section{Additional Evaluation Results}
\label{sec:appendix_eval}

\begin{figure*}[ht]
  \centering
   \includegraphics[width=\textwidth]{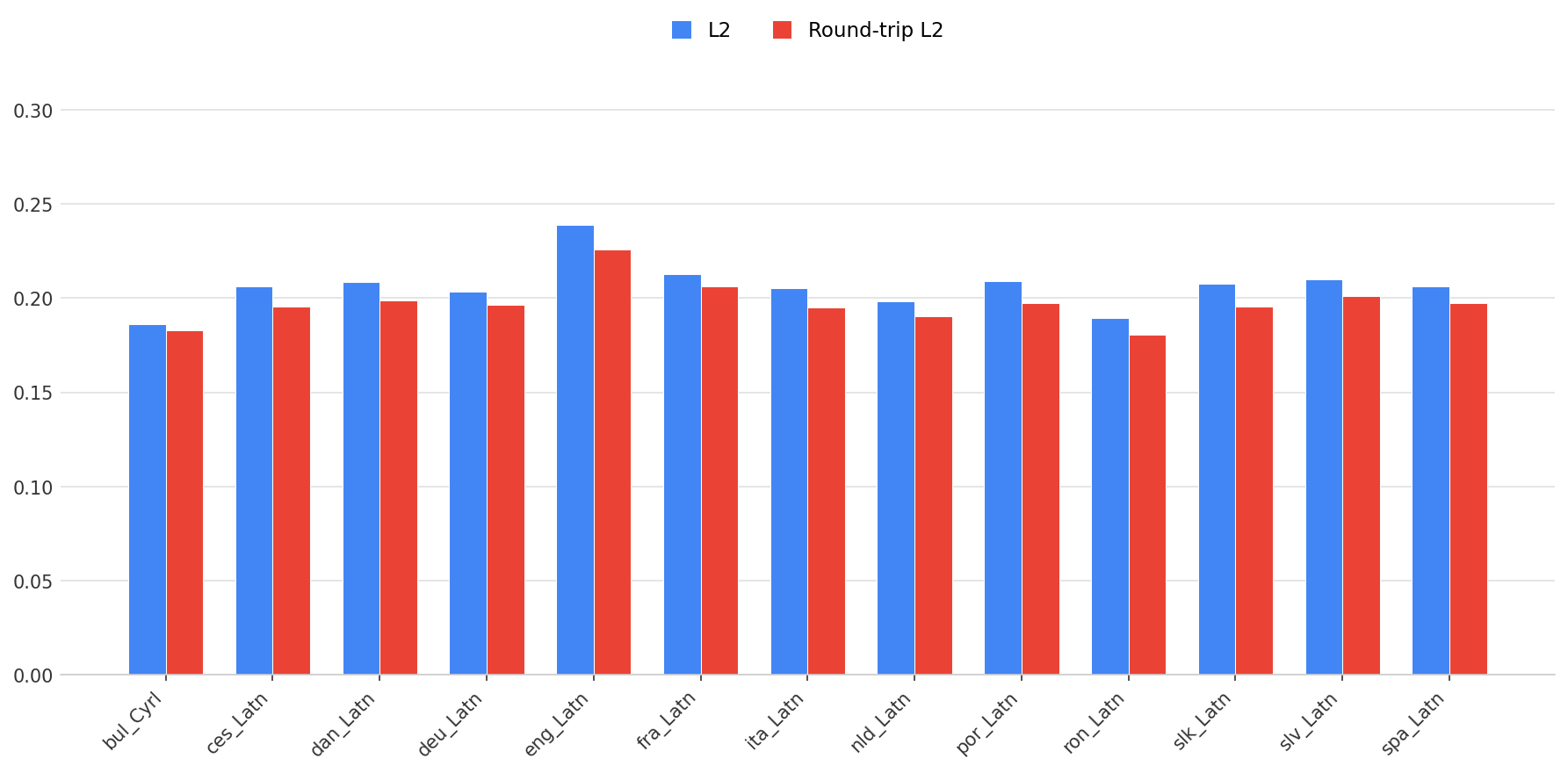}
   \caption{Bar chart showcasing pre-train evaluation results using L2 and Round-trip L2 for the MultiEURLEX dataset}
   \label{fig:eurlex_results}
\end{figure*}

\begin{figure*}[ht]
  \centering
   \includegraphics[width=\textwidth]{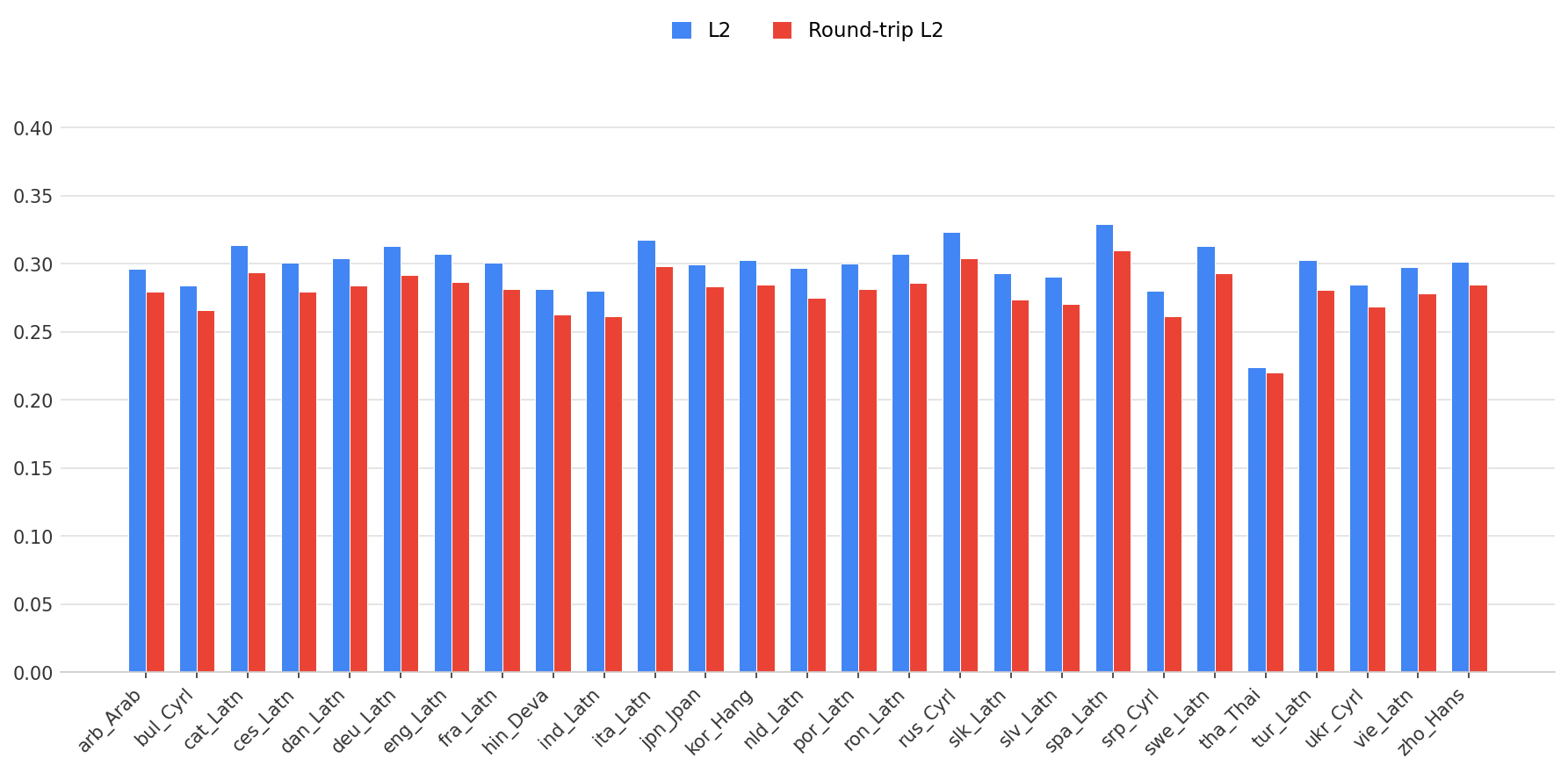}
   \caption{Bar chart showcasing pre-train evaluation results using L2 and Round-trip L2 for the Wiki40B dataset}
   \label{fig:wiki_results}
\end{figure*}

\begin{figure*}[ht]
  \centering
   \includegraphics[width=\textwidth]{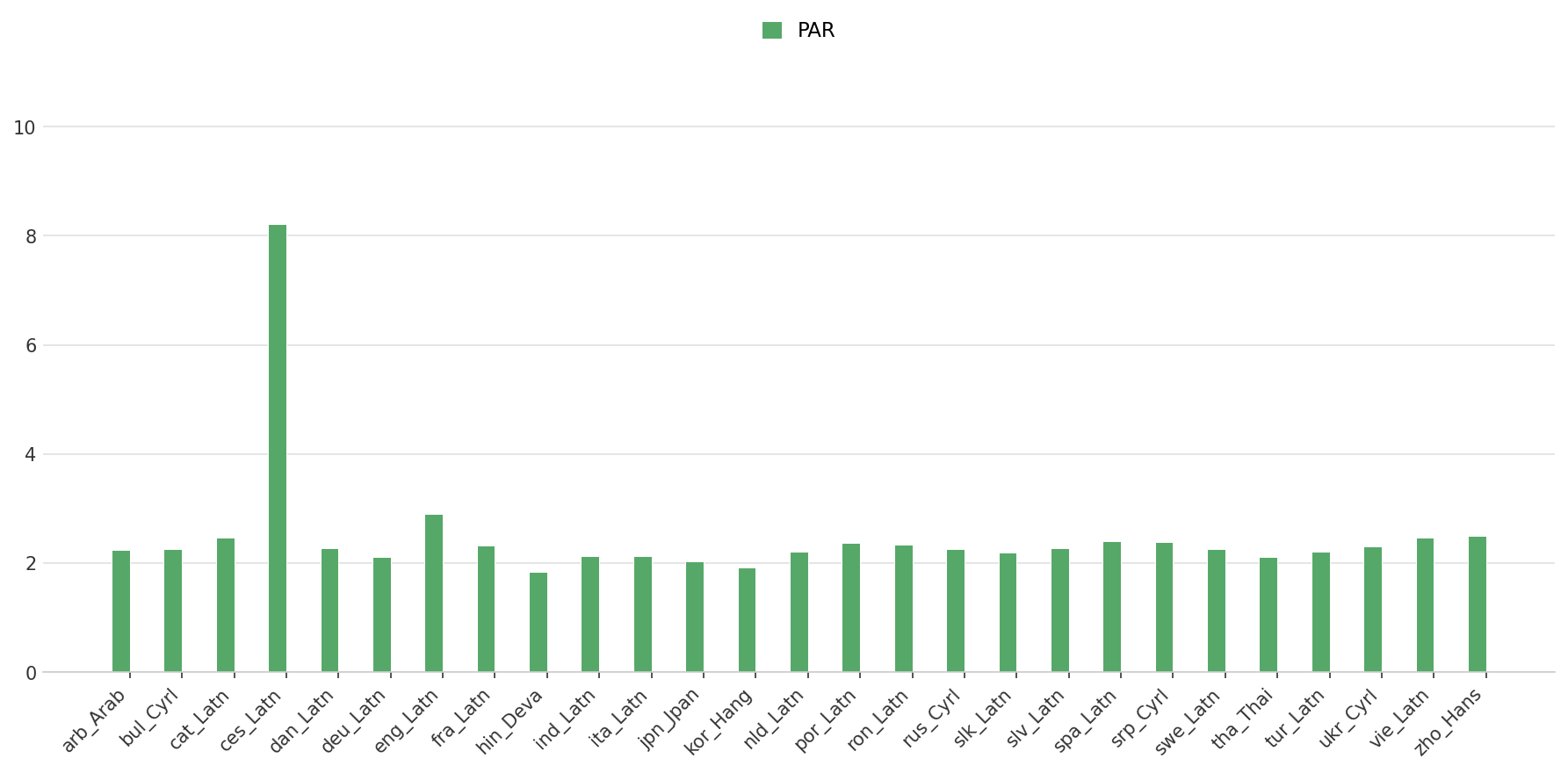}
   \caption{Bar chart showcasing pre-train evaluation results using PAR for the C4 dataset}
   \label{fig:c4_results_par}
\end{figure*}

\begin{figure*}[ht]
  \centering
   \includegraphics[width=\textwidth]{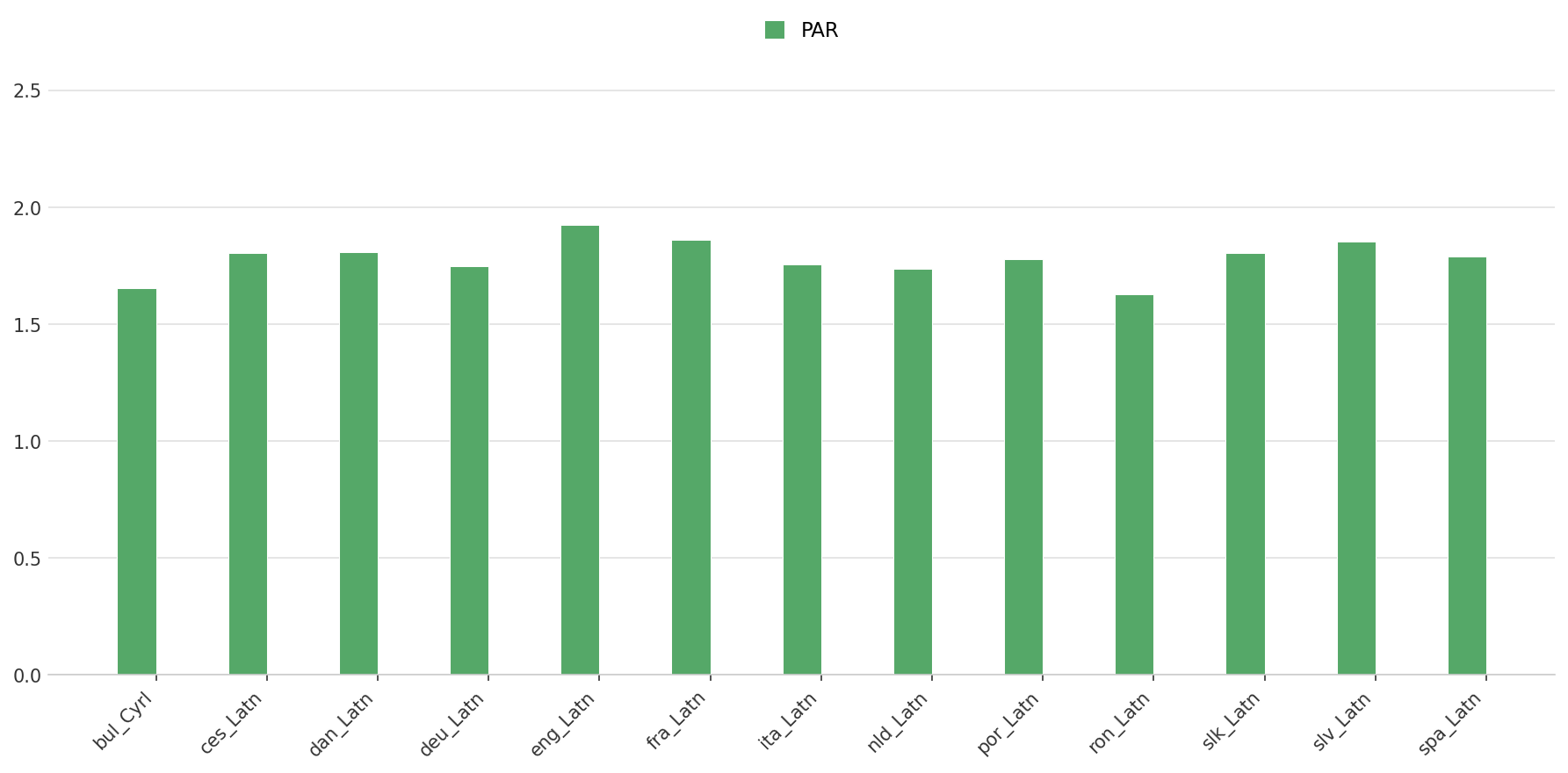}
   \caption{Bar chart showcasing pre-train evaluation results using PAR for the MultiEURLEX dataset}
   \label{fig:eurlex_results_par}
\end{figure*}

\begin{figure*}[ht]
  \centering
   \includegraphics[width=\textwidth]{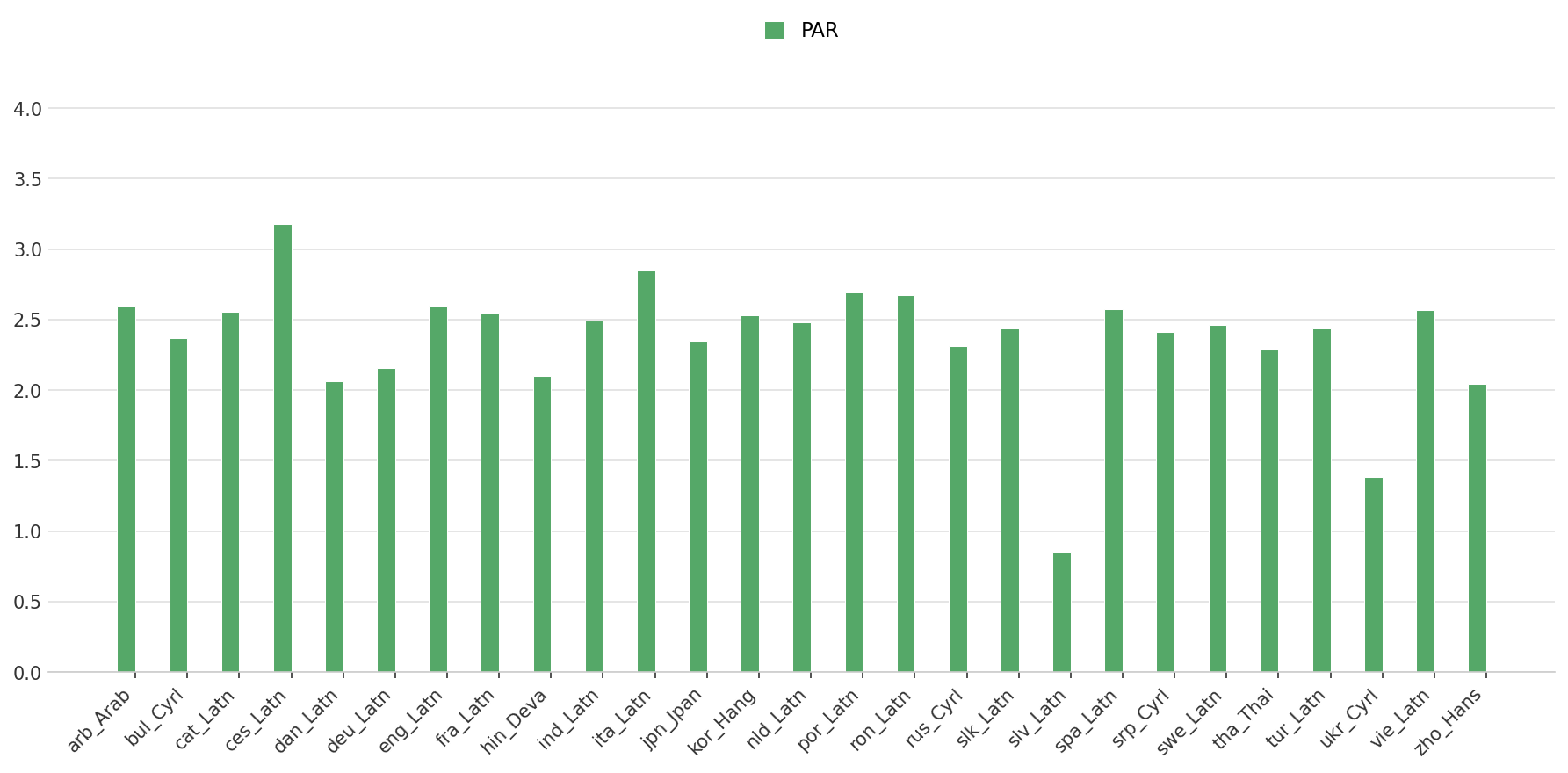}
   \caption{Bar chart showcasing pre-train evaluation results using PAR for the Wiki40B dataset}
   \label{fig:wiki_results_par}
\end{figure*}

\begin{figure*}[ht]
  \centering
   \includegraphics[width=\textwidth]{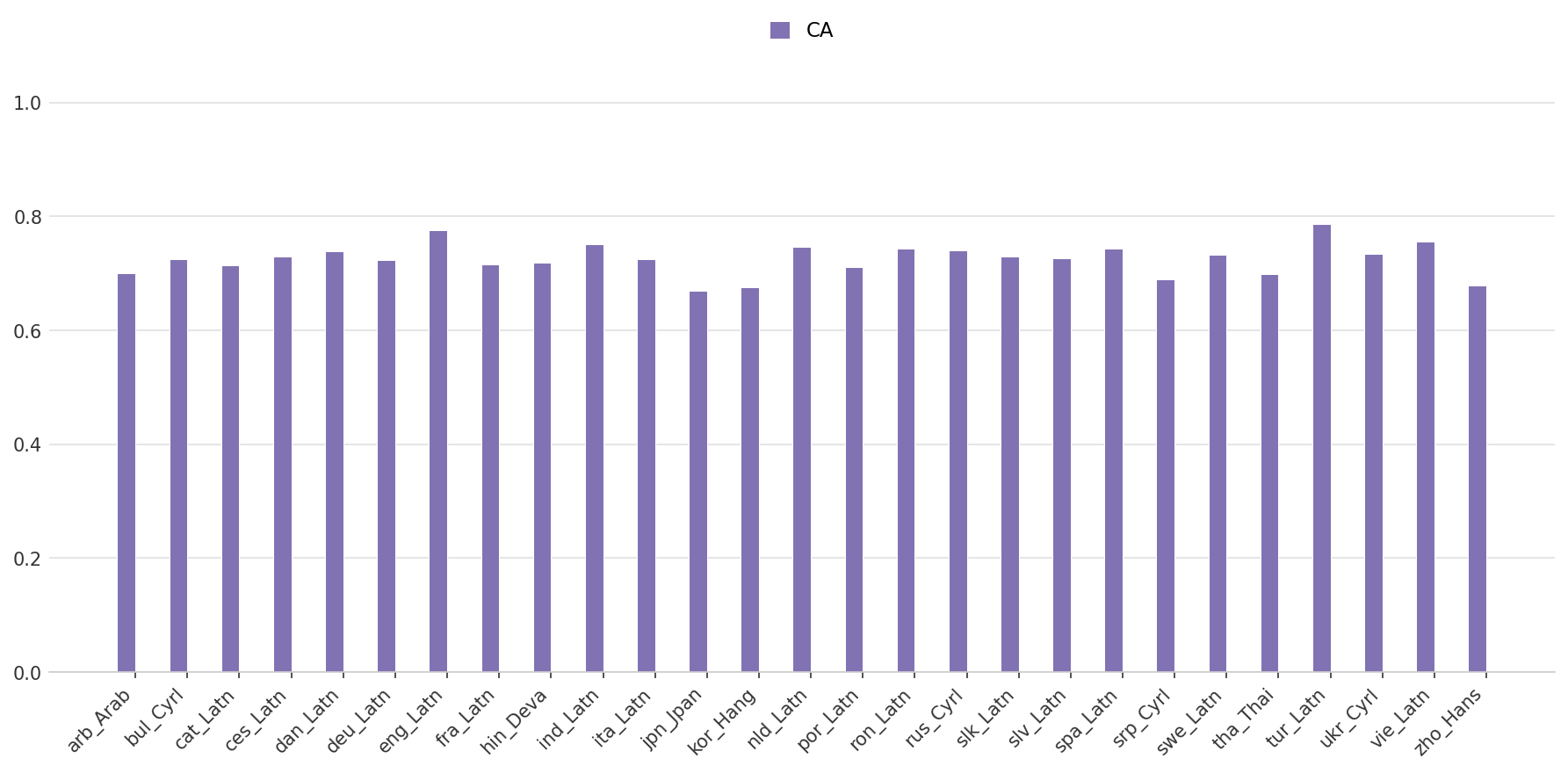}
   \caption{Bar chart showcasing pre-train evaluation results using CA for the C4 dataset}
   \label{fig:c4_results_ca}
\end{figure*}

\begin{figure*}[ht]
  \centering
   \includegraphics[width=\textwidth]{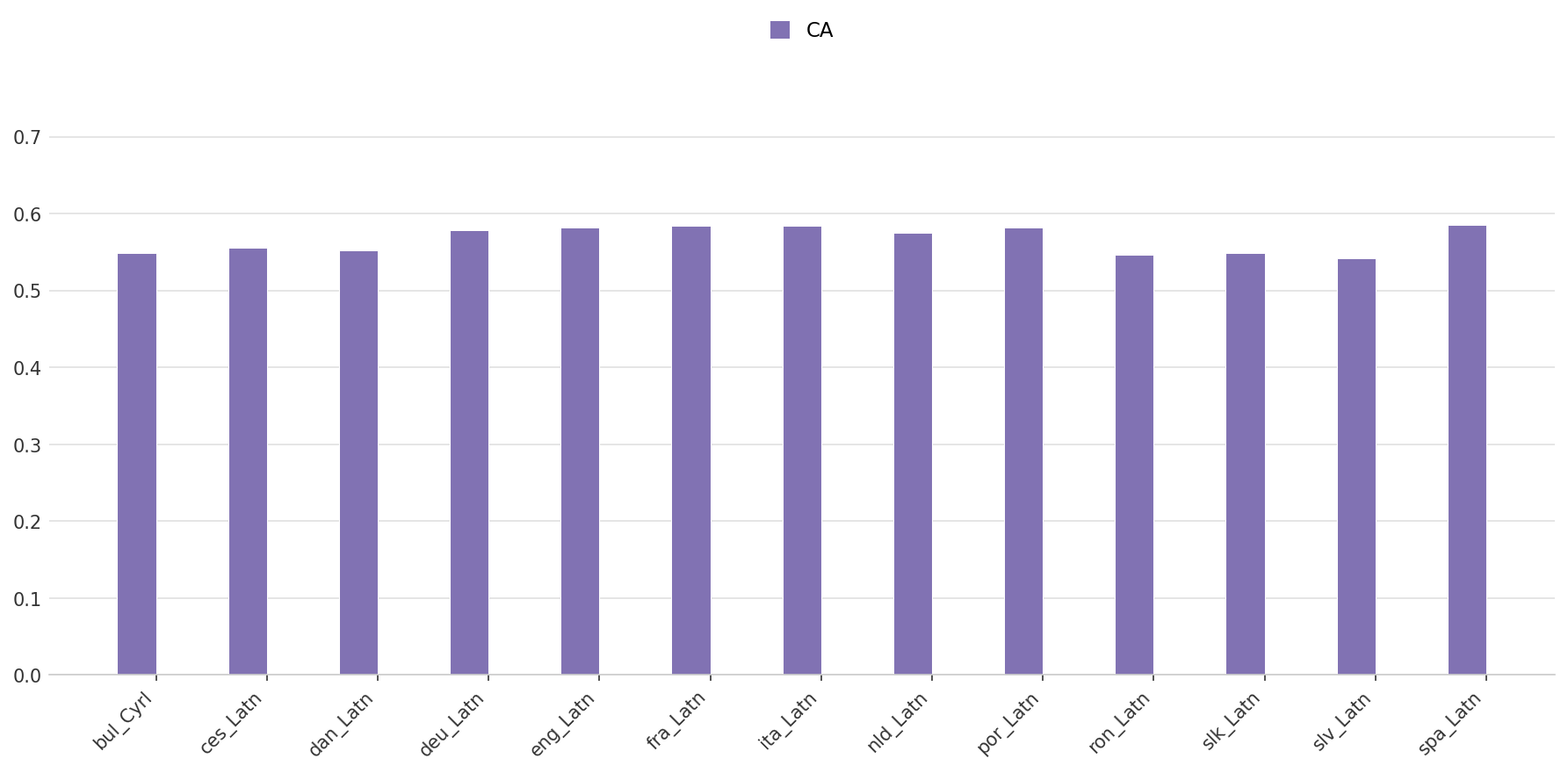}
   \caption{Bar chart showcasing pre-train evaluation results using CA for the MultiEURLEX dataset}
   \label{fig:eurlex_results_ca}
\end{figure*}

\begin{figure*}[ht]
  \centering
   \includegraphics[width=\textwidth]{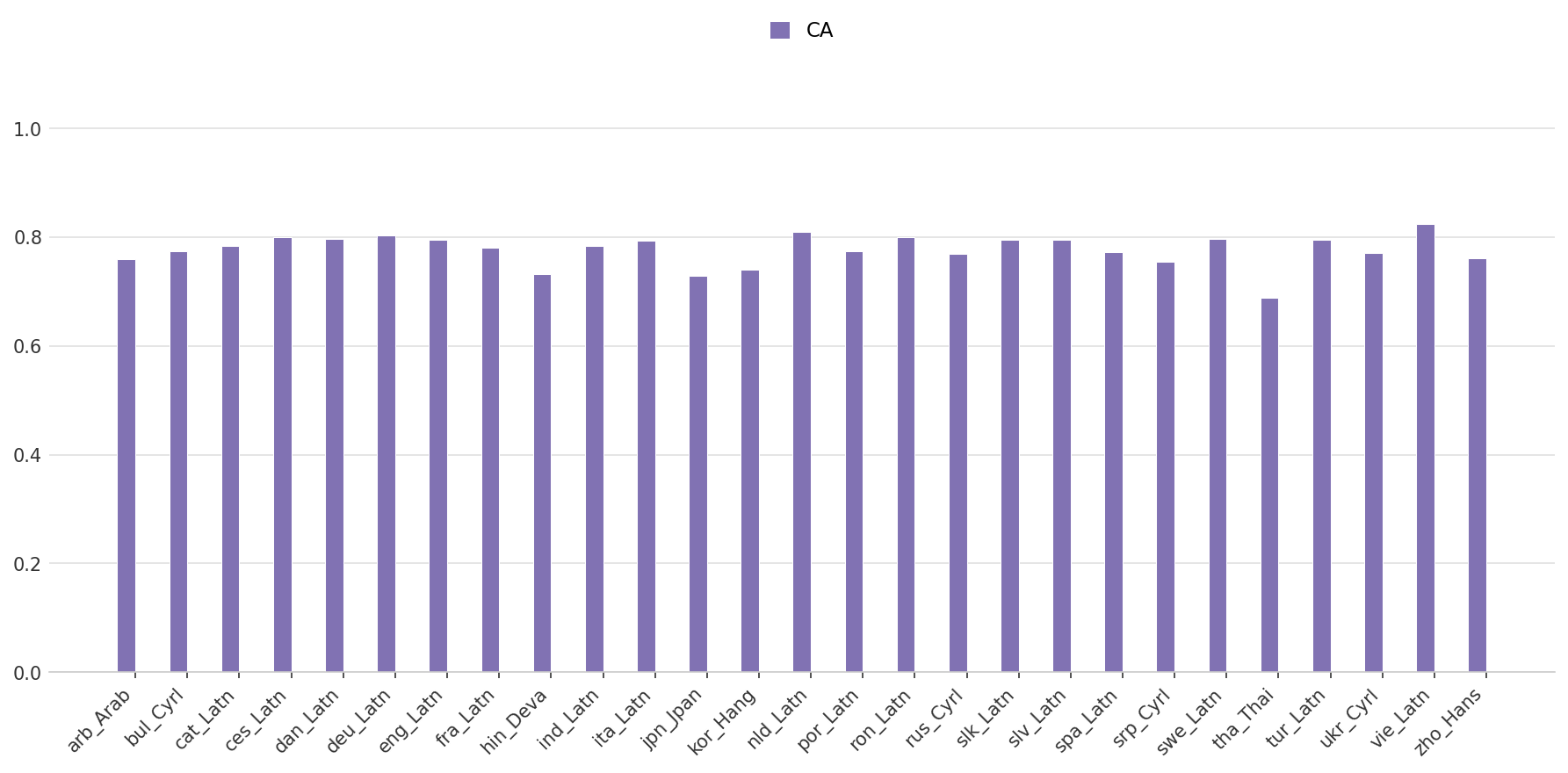}
   \caption{Bar chart showcasing pre-train evaluation results using CA for the Wiki40B dataset}
   \label{fig:wiki_results_ca}
\end{figure*}

Results for the pilot study of cosine similarity for English to target language SONAR embeddings are shown in \Cref{tab:cos_sim}.
We provide pre-training evaluation results for MultiEURLEX and Wiki40B split by language for L2 and R-L2 in \Cref{fig:eurlex_results} and in \Cref{fig:wiki_results}, respectively.
We report pre-training evaluation results for C4, MultiEURLEX and Wiki40B split by language for PAR in \Cref{fig:c4_results_par}, in \Cref{fig:eurlex_results_par} and in \Cref{fig:wiki_results_par}, respectively.
The pre-training evaluation results for C4, MultiEURLEX and Wiki40B split by language for CA are presented in \Cref{fig:c4_results_ca}, in \Cref{fig:eurlex_results_ca} and in \Cref{fig:wiki_results_ca}, respectively.

Interestingly, we observe that Mimir presented repetitive outputs for ``ces\_Latn'' on C4 as demonstrated by the spike for that language in \Cref{fig:c4_results_par}.

To evaluate Mimir's ability to understand context, we test the model on the Word Sense Disambiguation (WSD) task.
WSD is particularly relevant to Concept Modeling, as it requires precise contextual understanding to distinguish among different senses of the same word.

We perform the evaluation leveraging the XL-WSD \citep{pasini2021xl} benchmark extended for LLM evaluation by \citet{basile2025exploring}.
Specifically, we consider the generative split of the dataset without translations, comprising 6,757 instances in English, 1,673 in Italian, 1,248 in Spanish, 539 in French, and 263 in German.
As baselines, we report the zero-shot inference results for the Llama-3.1-8B-Instruct and Llama-3.1-405B-Instruct models, as reported by \citet{basile2025exploring}.
Following \citet{basile2025exploring}, we use ROUGE-L as the primary evaluation metric.
Results are reported in \Cref{tab:wsd_eval}.
Results show that Mimir generally underperforms both Llama 3.1 8B and 405B on this task. 
Nevertheless, it demonstrates remarkable multilingual performance, achieving non-trivial performance even on lower-resource evaluation languages such as German (14.42 ROUGE-L).
These findings suggest that while concept-based modeling alone is insufficient to match state-of-the-art token-based LLMs on WSD, multilingual concept representations can capture meaningful semantic information that could be further improved through targeted training for sense understanding.

\begin{figure*}[ht]
  \centering
   \includegraphics[width=0.9\textwidth]{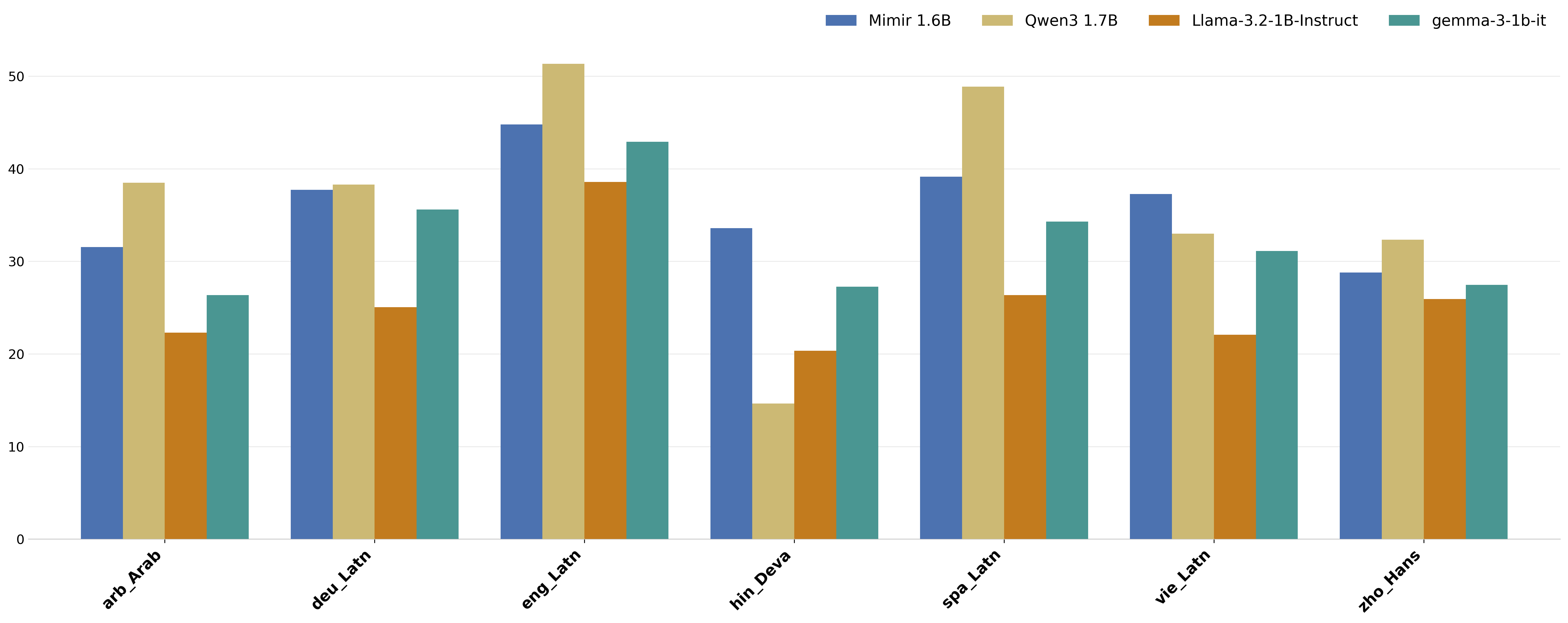}
   \caption{Bar chart showcasing instruction-tuning evaluation results using ROUGE-L for the MLQA dataset with additional baselines}
   \label{fig:mlqa_language_results_LLMs}
\end{figure*}

\begin{figure*}[ht]
  \centering
   \includegraphics[width=0.9\textwidth]{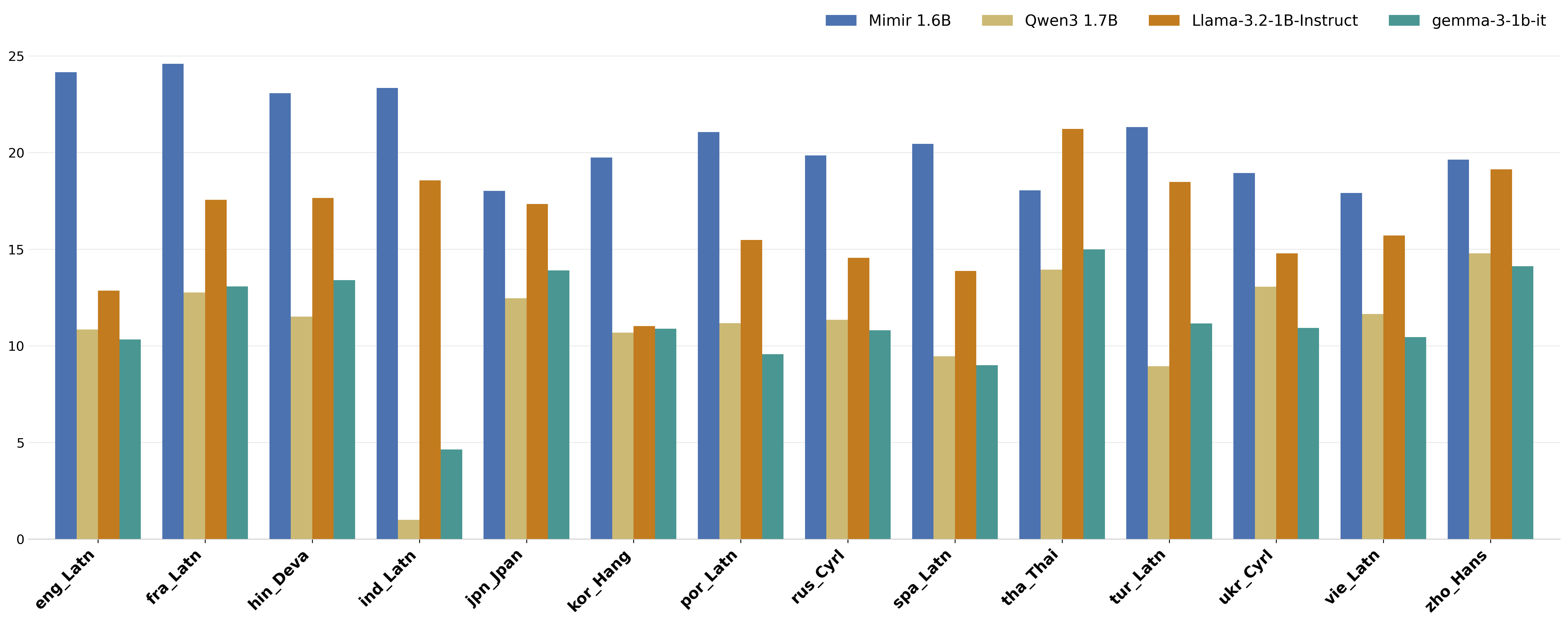}
   \caption{Bar chart showcasing instruction-tuning evaluation results using ROUGE-L for the XL-Sum dataset with additional baselines}
   \label{fig:xlsum_language_results_LLMs}
\end{figure*}

We present results with additional baselines, which are Llama-3.2-1B-Instruct \citep{grattafiori2024llama} and gemma-3-1b-it \citep{team2025gemma} in \Cref{fig:mlqa_language_results_LLMs} for MLQA and in \Cref{fig:xlsum_language_results_LLMs} for XL-Sum.

\section{Contamination Analysis}
\label{sec:appendix_contam}

\begin{table*}[h]
\centering
\begin{tabular}{lrr}
\toprule
Language & Context & Question \\
\midrule
arb\_Arab & 0.00\% & 0.00\% \\
deu\_Latn & 0.00\% & 0.02\% \\
eng\_Latn & 0.30\% & 0.11\% \\
hin\_Deva & 0.00\% & 0.02\% \\
spa\_Latn & 0.00\% & 0.06\% \\
vie\_Latn & 0.00\% & 0.00\% \\
zho\_Hans & 0.00\% & 0.02\% \\
\bottomrule
\end{tabular}
\caption{Contamination analysis for MLQA, reported at the context and question levels}
\label{tab:contam_mlqa}
\end{table*}

\begin{table*}[h]
\centering
\begin{tabular}{lrr}
\toprule
Language & Text & Summary \\
\midrule
eng\_Latn & 0.01\% & 0.26\% \\
fra\_Latn & 0.00\% & 0.09\% \\
hin\_Deva & 0.02\% & 0.64\% \\
ind\_Latn & 0.00\% & 0.08\% \\
jpn\_Jpan & 0.00\% & 0.11\% \\
kor\_Hang & 0.00\% & 0.55\% \\
por\_Latn & 0.08\% & 0.56\% \\
rus\_Cyrl & 0.15\% & 4.24\% \\
spa\_Latn & 0.02\% & 0.76\% \\
tha\_Thai & 0.00\% & 1.21\% \\
tur\_Latn & 0.00\% & 0.77\% \\
ukr\_Cyrl & 0.00\% & 0.81\% \\
vie\_Latn & 0.00\% & 0.45\% \\
zho\_Hans & 0.32\% & 2.51\% \\
\bottomrule
\end{tabular}
\caption{Contamination analysis for XL-Sum, reported at the text and summary levels}
\label{tab:contam_xlsum}
\end{table*}

We use the original train and test splits released for both MLQA and XL-Sum, we did not perform any splitting ourselves. 
For MLQA specifically, the training set is derived from SQuAD. 
To assess potential data contamination, we report a per-language analysis of contamination between the training and test splits for both datasets. 
We measure overlap (check if each test string is present in the train strings) using exact match after normalization (lowercasing, punctuation removal and collapsing consecutive whitespace characters into a single one). 
For MLQA, we check the context and question levels; for XL-Sum, the article text and summary levels. 
Complete results are reported in \Cref{tab:contam_mlqa} for MLQA and in \Cref{tab:contam_xlsum} for XL-Sum. 
Overlap is absent or very low across all languages (below 2\%); the only exceptions are the Russian and Chinese summaries in XL-Sum (4.24\% and 2.51\%). 
Hence, the observed gains are not driven by test split memorization. 

\section{Licenses}

We will release pre-train and instruction-tuning data under a license that complies with the original datasets present in the training mixture.
We will release the Mimir model and the codebase under MIT license.

\section{Use of AI Assistants}

During the preparation of this paper, we used Gemini 3.1 Pro to assist with code optimization of our scripts. 
Additionally, we used Grammarly to check grammar and improve the readability of the paper. 
We have reviewed all AI-optimized code and generated text, and we take full responsibility for the accuracy, originality, and final content of this work.

\begin{table*}[ht]
  \centering
  \begin{tabular}{@{}llll@{}}
    \toprule
    \textbf{Language} & \textbf{Instances} & \textbf{Sentences} \\
    \midrule
    arb\_Arab & 6,920,272 & 250,934,094 \\
    bel\_Cyrl & 234,579 & 9,458,247 \\ 
    ben\_Beng & 1,695,605 & 53,614,461 \\
    bos\_Latn & 2,371,974 & 68,947,657 \\
    bul\_Cyrl & 2,902,512 & 114,991,120 \\
    cat\_Latn & 1,913,415 & 49,630,459 \\
    ces\_Latn & 7,377,000 & 293,734,712 \\
    cym\_Latn & 92,886 & 2,985,904 \\
    dan\_Latn & 5,068,336 & 205,778,366 \\
    deu\_Latn & 55,378,296 & 2,290,435,652 \\
    eng\_Latn & 339,347,842 & 17,363,106,068 \\
    fra\_Latn & 40,203,392 & 1,528,389,036 \\
    heb\_Hebr & 1,618,116 & 72,240,173 \\
    hin\_Deva & 2,406,778 & 74,217,167 \\
    hrv\_Latn & 691,812 & 54,837,623 \\
    ind\_Latn & 11,192,416 & 479,630,974 \\
    ita\_Latn & 26,684,472 & 919,861,585 \\
    jpn\_Jpan & 44,678,584 & 2,300,431,291 \\
    kan\_Knda & 266,972 & 8,544,099 \\
    kor\_Hang & 6,797,096 & 301,721,036 \\
    lvs\_Latn & 896,648 & 57,697,945 \\
    mal\_Mlym & 370,986 & 13,204,579 \\
    mar\_Deva & 436,884 & 14,323,222 \\
    mkd\_Cyrl & 463,482 & 11,311,454 \\
    nld\_Latn & 16,447,336 & 566,173,272 \\
    npi\_Deva & 545,802 & 12,861,045 \\
    ory\_Orya & 144,953 & 2,969,207 \\
    pol\_Latn & 16,968,272 & 657,652,847 \\
    por\_Latn & 22,302,304 & 757,671,212 \\
    ron\_Latn & 6,510,056 & 240,522,807 \\
    rus\_Cyrl & 78,058,128 & 4,847,502,965 \\
    slk\_Latn & 3,348,784 & 134,859,963 \\
    slv\_Latn & 1,346,495 & 70,469,351 \\
    spa\_Latn & 49,273,160 & 1,602,905,444 \\
    srp\_Cyrl & 462,948 & 20,646,830 \\
    swe\_Latn & 6,642,000 & 241,586,555 \\
    swh\_Latn & 134,693 & 4,211,019 \\
    tam\_Taml & 617,340 & 23,817,904 \\
    tel\_Telu & 219,340 & 11,473,008 \\
    tha\_Thai & 4,008,200 & 235,314,468 \\
    tur\_Latn & 10,621,912 & 390,860,061 \\
    ukr\_Cyrl & 5,929,224 & 227,975,182 \\
    urd\_Arab & 537,024 & 15,927,173 \\
    vie\_Latn & 6,818,296 & 316,157,489 \\
    zho\_Hans & 51,449,701 & 1,665,507,651 \\
    zho\_Hant & 9,049,706 & 296,894,863 \\
    \bottomrule
  \end{tabular}
  \caption{Complete list of all languages included in pre-training and their cardinality (both number of instances and number of sentences)}
  \label{tab:dataset}
\end{table*}

\begin{table*}[ht]
  \centering
  \begin{tabular}{@{}lllllll@{}}
    \toprule
    \textbf{Language} & \textbf{Aya} & \textbf{Bactrian-X} & \textbf{Open Assistant 2} & \textbf{Ours} & \textbf{MLQA} & \textbf{XL-Sum} \\
    \midrule
    arb\_Arab & 66,924 & 794,330 & 120 & 16,210 & 968,702 & 0 \\
    ben\_Beng & 21,709 & 902,598 & 0 & 3,922 & 0 & 0 \\
    ces\_Latn & 0 & 798,389 & 15 & 20,178 & 0 & 0 \\
    deu\_Latn & 2,451 & 843,549 & 20,244 & 167,369 & 1,022,780 & 0 \\
    eng\_Latn & 46,309 & 853,400 & 236,677 & 1,510,862 & 1,022,762 & 9,218,354\\
    fra\_Latn & 13,504 & 825,812 & 10,298 & 112,023 & 0 & 264,330\\
    heb\_Hebr & 0 & 769,294 & 29 & 4,692 & 0 & 0 \\
    hin\_Deva & 12,444 & 860,428 & 0 & 6,964 & 1,041,678 & 2,569,314 \\
    hrv\_Latn & 0 & 831,293 & 0 & 2,306 & 0 & 0 \\
    ind\_Latn & 11,177 & 880,484 & 31 & 30,313 & 0 & 1,182,739 \\
    ita\_Latn & 6,176 & 796,651 & 2,643 & 76,201 & 0 & 0 \\
    jpn\_Jpan & 61,121 & 845,659 & 1,787 & 101,772 & 0 & 251,813 \\
    kor\_Hang & 3,866 & 864,464 & 25 & 20,903 & 0 & 165,207 \\
    lvs\_Latn & 0 & 816,503 & 0 & 1,913 & 0 & 0 \\
    mal\_Mlym & 17,462 & 893,154 & 0 & 2,495 & 0 & 0 \\
    mar\_Deva & 34,961 & 872,085 & 0 & 1,245 & 0 & 0 \\
    mkd\_Cyrl & 0 & 793,209 & 0 & 2,522 & 0 & 0 \\
    nld\_Latn & 20,139 & 835,846 & 144 & 46,460 & 0 & 0 \\
    npi\_Deva & 36,699 & 898,382 & 0 & 2,528 & 0 & 0\\
    pol\_Latn & 13,080 & 808,567 & 900 & 50,124 & 0 & 0 \\
    por\_Latn & 61,850 & 808,277 & 12,656 & 61,529 & 0 & 2,134,442 \\
    ron\_Latn & 0 & 810,049 & 0 & 19,301 & 0 & 0 \\
    rus\_Cyrl & 4,034 & 785,054 & 48,847 & 224,329 & 0 & 2,576,458 \\
    slv\_Latn & 0 & 834,416 & 0 & 3,258 & 0 & 0 \\
    spa\_Latn & 32,801 & 824,865 & 106,204 & 138,844 & 1,204,589 & 1,863,457 \\
    swe\_Latn & 10,717 & 801,663 & 0 & 17,772 & 0 & 0 \\
    swh\_Latn & 3,045 & 860,170 & 0 & 2,382 & 0 & 0 \\
    tam\_Taml & 293,226 & 859,460 & 0 & 1,779 & 0 & 0 \\
    tel\_Telu & 160,897 & 799,914 & 0 & 1,714 & 0 & 0 \\
    tha\_Thai & 6,988 & 971,516 & 4,150 & 16,106 & 0 & 392,155 \\
    tur\_Latn & 33,122 & 850,842 & 99 & 28,978 & 0 & 893,077\\
    ukr\_Cyrl & 5,655 & 784,703 & 3,061 & 18,235 & 0 & 1,468,201 \\
    urd\_Arab & 24,810 & 849,140 & 0 & 2,357 & 0 & 0 \\
    vie\_Latn & 171,693 & 838,376 & 365 & 17,882 & 1,032,551 & 1,503,432 \\
    zho\_Hans & 21,897 & 719,010 & 20,022 & 142,966 & 1,204,423 & 1,108,904 \\
    \bottomrule
  \end{tabular}
  \caption{Complete list of all languages included in instruction tuning, the source dataset and their cardinality (number of sentences)}
  \label{tab:dataset_it}
\end{table*}

\begin{table*}[ht]
    \begin{tabular}{p{\linewidth}}
    \toprule
    \rowcolor{blue!15}
    CONTEXT\\
    \rowcolor{gray!15}
    \{USER\_TURN\_SENTENCE\}\\
    \rowcolor{gray!15}
    \{PROMPT\_SENTENCES\}\\
    \rowcolor{gray!15}
    \{ASSISTANT\_TURN\_SENTENCE\}\\
    \rowcolor{blue!15}
    COMPLETION\\
    \rowcolor{gray!15}
    \{RESPONSE\_SENTENCES\}\\
    \rowcolor{gray!15}
    End of text.\\
    \bottomrule
    \end{tabular}
    \caption{Prompt used during instruction tuning. \{USER\_TURN\_SENTENCE\} is the ``User turn.'' sentence translated to the language of the conversation. \{ASSISTANT\_TURN\_SENTENCE\} is the ``Assistant turn.'' sentence translated to the language of the conversation. \{PROMPT\_SENTENCES\} is the list of sentences obtained from the prompt using a sentence splitting model. \{RESPONSE\_SENTENCES\} is the list of sentences obtained from the assistant response using a sentence splitting model.}
    \label{tab:prompt}
\end{table*}

\begin{table*}[ht]
    \footnotesize
    \begin{tabular}{p{\linewidth}}
    \toprule
    \rowcolor{gray!15}
    You are an expert synthetic data generator. Your task is to generate a realistic, multi-turn conversation between a USER and an AI ASSISTANT based on a specific topic. \newline
    \newline
    \textbf{SETTINGS:} \newline
    - \textbf{Topic:} \{TOPIC\} \newline
    - \textbf{Target Language:} \{TGT\_LANG\} \newline
    - \textbf{Conversation Length:} Approximately \{NUM\_TURNS\} turns (total messages between User and Assistant). \newline
    \newline
    \textbf{INSTRUCTIONS FOR ``USER'' GENERATION:} \newline
    1. \textbf{Language:} Fluent, natural \{TGT\_LANG\}. \newline
    2. \textbf{Content:} Start by asking about the Topic. Follow-up questions should dig deeper, asking for opinions, comparisons, or specific details. \newline
    3. \textbf{Cultural Relevance:} The User's perspective, idioms, and context must be culturally relevant to \{TGT\_LANG\}. (e.g., if the language is Japanese and the topic is `Lunch', discuss Bento or Ramen, not PB\&J sandwiches). \newline
    \newline
    \textcolor{red}{\textbf{USER CONSTRAINTS:}} \newline
    \textcolor{red}{- \textbf{Length Constraints:} Approximately 20\% of the time, the User must explicitly constrain the Assistant's output length (e.g., `Answer with a single word', `Give me a bulleted list', `Keep it under 10 words').} \newline
    \textcolor{red}{- \textbf{Unconventional Formatting:}  Occasionally, the User must request the response in an unusual format or persona (e.g., 'Return the answer as a valid JSON dictionary', 'Answer in pirate speech').} \newline
    \newline
    \textbf{INSTRUCTIONS FOR ``ASSISTANT'' GENERATION:} \newline
    1. \textbf{Language:} Fluent, natural \{TGT\_LANG\}. \newline
    2. \textbf{Behavior:} Helpful, accurate, and culturally aware, unless a specific interaction scenario (below) requires otherwise. \newline
    3. \textbf{Responsiveness:} If the User sets a length or format constraint, the Assistant MUST strictly obey it. \newline
    \newline
    \textcolor{red}{\textbf{REQUIRED INTERACTION SCENARIOS:}} \newline
    \textcolor{red}{Select AT LEAST ONE of the following specific scenarios to naturally weave into the conversation:} \newline
    \newline
    \textcolor{red}{\textbf{NEGATIVE/ADVERSARIAL SCENARIOS:}} \newline
    \textcolor{red}{1. \textbf{The `Hallucination' \& Correction:} The ASSISTANT provides a factually incorrect answer. The USER corrects it. The ASSISTANT apologizes and provides the correct answer.} \newline
    \textcolor{red}{2. \textbf{The Valid Refusal (Impossible Request):} The USER asks an impossible or out-of-bounds question. The ASSISTANT politely declines to answer.} \newline
    \textcolor{red}{3. \textbf{The False Correction (Assistant stands its ground):} The ASSISTANT provides a correct answer. The USER incorrectly claims it is wrong (and may provide a wrong alternative). The ASSISTANT politely but firmly asserts its correctness and explains why.} \newline
    \newline
    \textcolor{red}{\textbf{COMPLEX INTERACTION SCENARIOS:}} \newline
    \textcolor{red}{1. \textbf{The Ambiguous Query:} The USER asks a vague follow-up using unclear pronouns (e.g., `What about that other one?'). The ASSISTANT must politely ask for clarification before answering.} \newline
    \textcolor{red}{2. \textbf{The Compound Question:} The USER asks at least 3 distinct questions in a single message. The ASSISTANT must systematically answer all parts.} \newline
    \textcolor{red}{3. \textbf{The Pivot:} The USER abruptly changes the sub-topic mid-conversation. The ASSISTANT follows the pivot smoothly.} \newline
    \textcolor{red}{4. \textbf{The Goalpost Move:} The USER asks the ASSISTANT to rewrite its previous answer with a completely new constraint (e.g., `Now make it rhyme' or `Explain it like I am 5').} \newline
    \newline
    \textbf{OUTPUT FORMAT:} \newline
    Return strictly a valid JSON list of dictionaries. Do not include markdown formatting (like \texttt{
```json}). \newline
    Format: \newline
    [ \newline
      \{"turn": 1, "role": "user", "content": "..."\}, \newline
      \{"turn": 1, "role": "assistant", "content": "..."\}, \newline
      ... \newline
    ] \\
    \bottomrule
    \end{tabular}
    \caption{Prompt used for synthetic generation of multilingual multi-turn conversational data. \{TOPIC\} is a placeholder for the complete topic sampled from the Everyday Conversations dataset. \{TGT\_LANG\} is a placeholder for the target language and script (e.g. ``fra\_Latn''). \{NUM\_TURNS\} is a placeholder for a random number of turns selected randomly in [4, 6, 8, 10, 12]. Text in \textcolor{red}{red} is randomly removed from the prompt.}
    \label{tab:prompt_synth}
\end{table*}

\begin{table*}[h]
    \footnotesize
    \begin{tabular}{p{\linewidth}}
    \toprule
    \rowcolor{gray!15}
    You are an expert synthetic data generator. Your task is to generate a realistic, multi-turn conversation between a USER and an AI ASSISTANT focused on solving and discussing a specific math problem. \newline
    \newline
    \textbf{REFERENCE MATERIAL (ENGLISH):} \newline
    - \textbf{Math Problem:} \{PROBLEM\} \newline
    - \textbf{Reference Solution:} \{SOLUTION\} \newline \newline
    \textbf{SETTINGS:} \newline
    - \textbf{Target Language:} \{TGT\_LANG\} \newline
    - \textbf{Conversation Length:} Approximately \{NUM\_TURNS\} turns (total messages between User and Assistant). \newline
    \newline
    \textbf{INSTRUCTIONS FOR ``USER'' GENERATION:} \newline
    1. \textbf{Language \& Natural Framing:} Fluent, natural \{TGT\_LANG\}. Do NOT directly or stiffly translate the English reference problem. Internalize the problem, and have the User ask it naturally in \{TGT\_LANG\} as if they just encountered it in their homework or daily life. \newline
    2. \textbf{Progression:} Start by presenting the math problem. Follow-up questions should dig deeper into the methodology, ask for alternative ways to solve it, or introduce the complex/adversarial scenarios below. \newline
    \textcolor{red}{\textbf{USER CONSTRAINTS:}} \newline
    \textcolor{red}{- \textbf{Direct Answer Constraint:} Approximately 20\% of the time, the User's first message must explicitly ask for the final answer ONLY, without any reasoning or step-by-step breakdown (e.g., `Just give me the final number', `Answer directly without explanation').} \newline
    \textcolor{red}{- \textbf{Unconventional Formatting:} Occasionally, the User must request the math steps in a specific format (e.g., 'Return the steps as a JSON list', 'Explain the logic using pirate speech', 'Put every mathematical operation in a separate bullet point').} \newline
    \newline
    \textbf{INSTRUCTIONS FOR ``ASSISTANT'' GENERATION:} \newline
    1. \textbf{Language:} Fluent, natural \{TGT\_LANG\}. Ensure mathematical terms are correctly translated. \newline
    2. \textbf{Accuracy:} The math must be flawlessly executed and align with the logic of the Reference Solution, unless a negative scenario explicitly requires an error. \newline
    3. \textbf{Responsiveness:} If the User asks for a direct answer without reasoning in their first turn, the Assistant MUST output exactly the final number/solution and nothing else. The Assistant must strictly obey all other length or formatting constraints requested by the User. \newline
    \newline
    \textcolor{red}{\textbf{REQUIRED INTERACTION SCENARIOS:}} \newline
    \textcolor{red}{Select AT LEAST ONE of the following specific scenarios to naturally weave into the conversation:} \newline
    \newline
    \textcolor{red}{\textbf{NEGATIVE/ADVERSARIAL MATH SCENARIOS:}} \newline
    \textcolor{red}{1. \textbf{The Calculation Error \& Correction:} The ASSISTANT makes a subtle calculation error or uses the wrong formula in one of the steps. The USER catches the math error and corrects it. The ASSISTANT apologizes, recalculates, and provides the correct answer.} \newline
    \textcolor{red}{2. \textbf{The Missing Information (Valid Refusal):} The USER asks a follow-up math question that lacks the necessary variables to be solved. The ASSISTANT politely explains what information is missing and why the calculation cannot be performed.} \newline
    \textcolor{red}{3. \textbf{The False Correction (Assistant stands its ground):} The ASSISTANT solves a step correctly. The USER incorrectly claims it is wrong based on a common math misconception (e.g., messing up the order of operations). The ASSISTANT politely but firmly asserts its correctness and explains the mathematical rule.} \newline
    \newline
    \textcolor{red}{\textbf{COMPLEX MATH INTERACTION SCENARIOS:}} \newline
    \textcolor{red}{1. \textbf{The Ambiguous Step Query:} The USER asks a vague follow-up about a specific number (e.g., `Where did that 3 come from?' or `Why did you multiply those two?'). The ASSISTANT must clarify the specific step in the reference solution.} \newline
    \textcolor{red}{2. \textbf{The Method Request:} The USER asks if there is an alternative mathematical way or formula to solve the exact same problem. The ASSISTANT provides a valid alternative method that yields the same result.} \newline
    \textcolor{red}{3. \textbf{The Variable Change (Goalpost Move):} Mid-conversation, the USER changes the numbers in the original problem (e.g., `What if there were 10 people instead of 5?'). The ASSISTANT recalculates everything based on the new parameters.} \newline
    \textcolor{red}{4. \textbf{The Concept Pivot:} The USER abruptly asks for the definition of a mathematical concept related to the problem (e.g., `By the way, what exactly is a factorial?'). The ASSISTANT explains it clearly, then ties it back to the current problem if applicable.} \newline
    \newline
    \textbf{OUTPUT FORMAT:} \newline
    Return strictly a valid JSON list of dictionaries. Do not include markdown formatting (like \texttt{```json}). \newline
    Format: \newline
    [ \newline
      \{"turn": 1, "role": "user", "content": "..."\}, \newline
      \{"turn": 1, "role": "assistant", "content": "..."\}, \newline
      ... \newline
    ] \\
    \bottomrule
    \end{tabular}
    \caption{Prompt used for synthetic generation of multilingual multi-turn math problems. \{PROBLEM\} and \{SOLUTION\} are placeholders for the reference problem and solution from the OpenMathInstruct-2 dataset. \{TGT\_LANG\} is a placeholder for the target language and script (e.g. ``fra\_Latn''). \{NUM\_TURNS\} is a placeholder for a random number of turns selected randomly in [4, 6, 8, 10, 12]. Text in \textcolor{red}{red} is randomly removed from the prompt.}
    \label{tab:prompt_synth_math}
\end{table*}

\begin{table*}[ht]
  \centering
  \begin{tabular}{@{}llll@{}}
    \toprule
    \textbf{Target Language} & \textbf{Cosine Similarity} \\
    \midrule
    arb\_Arab & 0.8208 \\
    bul\_Cyrl & 0.8447 \\
    cat\_Latn & 0.8213 \\
    ces\_Latn & 0.8071 \\
    dan\_Latn & 0.8643 \\
    deu\_Latn & 0.7642 \\
    ell\_Grek & 0.8701 \\
    est\_Latn & 0.8359 \\
    fin\_Latn & 0.7954 \\
    fra\_Latn & 0.8652 \\
    glg\_Latn & 0.8447 \\
    heb\_Hebr & 0.8169 \\
    hin\_Deva & 0.7617 \\
    hrv\_Latn & 0.8340 \\
    hun\_Latn & 0.7896 \\
    ind\_Latn & 0.8242 \\
    ita\_Latn & 0.8618 \\
    jpn\_Jpan & 0.6030 \\
    kat\_Geor & 0.5483 \\
    kor\_Hang & 0.5586 \\
    lit\_Latn & 0.8081 \\
    mar\_Deva & 0.4885 \\
    mkd\_Cyrl & 0.8589 \\
    nld\_Latn & 0.8657 \\
    pes\_Arab & 0.6865 \\
    pol\_Latn & 0.7925 \\
    ron\_Latn & 0.8599 \\
    rus\_Cyrl & 0.7905 \\
    slk\_Latn & 0.8394 \\
    slv\_Latn & 0.8481 \\
    spa\_Latn & 0.8628 \\
    srp\_Cyrl & 0.7856 \\
    swe\_Latn & 0.8125 \\
    tur\_Latn & 0.7510 \\
    ukr\_Cyrl & 0.7827 \\
    vie\_Latn & 0.7896 \\
    zho\_Hans & 0.6699 \\
    \bottomrule
  \end{tabular}
  \caption{Average cosine similarity for eng\_Latn to target language for SONAR embeddings}
  \label{tab:cos_sim}
\end{table*}

\begin{table*}[htb]
  \centering
  \large
  \begin{tabular}{@{}lll@{}}
    \toprule
    \textbf{Language} & \textbf{Model} & \textbf{ROUGE-L} \\
    \midrule
    \multirow{3}{*}{English} & Mimir 1.6B & 8.56 \\
    & Llama-3.1-8B-Instruct & 22.60 \\
    & Llama-3.1-405B-Instruct & 23.93 \\
    \midrule
    \multirow{3}{*}{French} & Mimir 1.6B & 15.82 \\
    & Llama-3.1-8B-Instruct & 19.01 \\
    & Llama-3.1-405B-Instruct & 17.51 \\
    \midrule
    \multirow{3}{*}{German} & Mimir 1.6B & 14.42 \\
    & Llama-3.1-8B-Instruct & 15.86 \\
    & Llama-3.1-405B-Instruct & 13.43 \\
    \midrule
    \multirow{3}{*}{Italian} & Mimir 1.6B & 10.37 \\
    & Llama-3.1-8B-Instruct & 13.63 \\
    & Llama-3.1-405B-Instruct & 15.24 \\
    \midrule
    \multirow{3}{*}{Spanish} & Mimir 1.6B & 15.06 \\
    & Llama-3.1-8B-Instruct & 18.11 \\
    & Llama-3.1-405B-Instruct & 19.15 \\
    \bottomrule
  \end{tabular}
  \caption{Results for evaluation on XL-WSD}
  \label{tab:wsd_eval}
\end{table*}

\clearpage
\clearpage
\clearpage
\clearpage

\bibliographystyle{unsrtnat}
\bibliography{references}  






\end{document}